\documentclass{article}
\usepackage{ijcai26}

\usepackage{times}
\usepackage{soul}
\usepackage{url}
\usepackage[hidelinks]{hyperref}
\usepackage[utf8]{inputenc}
\usepackage[small]{caption}
\usepackage{subcaption}
\usepackage{graphicx}
\usepackage{amsmath}
\usepackage{amssymb}
\usepackage{amsthm}
\usepackage{booktabs}
\usepackage{natbib}
\usepackage{listings}
\usepackage[dvipsnames]{xcolor}
\usepackage{algorithm, float}
\usepackage{algpseudocode}
\usepackage{todonotes}
\usepackage{multirow}
\usepackage{subfiles}

\definecolor{mygreen}{rgb}{0,0.6,0}
\definecolor{darkGreen}{rgb}{0,0.4,0}

\title{Using Weisfeiler-Leman Features for Algorithm Selection in Constraint Optimisation}

\author{
    Alessio Pellegrino\and
    Jacopo Mauro
    \affiliations
    University of Southern Denmark, Denmark
    \emails
    \{alessio, mauro\}@imada.sdu.dk
}

\begin{document}

\maketitle


\begin{abstract}
Algorithm Selection is essential for efficient Constraint Programming. Over the years, many algorithm selectors based on machine learning methods have been successfully applied, yet traditional feature extraction methods often rely on manually decided instance-level statistics that fail to capture the underlying problem structure.

In this paper we aim to bridge this gap by introducing a novel, automated feature extraction methodology that integrates graph conversion and Weisfeiler-Lehman graph kernels to generate robust structural representations of problem instances. The 1-WL test bounds the graph-distinguishing power of standard message-passing Graph Neural Networks (GNNs), and suitable GNN architectures match this bound \citep{Xuetal2018}. WL-based features offer an alternative that does not require training a GNN. Our primary contribution is a cut-based representation (\texttt{WLc}) designed to model structural partitions and provide a more nuanced predictive signal. We evaluate our approach on instances from the 2023--2025 MiniZinc Challenges across two tasks: maximizing Borda count scores and maximizing predictive accuracy. Experimental results across Support Vector Machines, Random Forests, and Multi-Layer Perceptrons demonstrate that cut-based features outperform \texttt{fzn2feat} with SVMs, while results with RFs and MLPs are closer.
\end{abstract}

\section{Introduction}
Algorithm Selection, the process of selecting the most appropriate algorithm from a portfolio for a given problem instance, is a critical component in the efficient execution of Constraint Programming models, as solver performance varies significantly across different problem instances. Traditional feature extraction methodologies, such as \texttt{fzn2feat}, primarily rely on instance-level statistics that may fail to capture the underlying latent structure or the complex interconnections between variables. While graph-based representations are designed to capture these structural nuances, they are typically coupled with deep neural networks that require substantial computational overhead. This creates a representational gap: non-neural machine learning techniques remain largely ``structure-blind,'' while structure-aware models are often too expensive for practical deployment.

This paper proposes a novel methodology to bridge this gap by implementing a graph conversion algorithm compatible with both classical and deep machine learning paradigms. Our approach utilizes Weisfeiler-Lehman graph kernels to generate robust representations of problem instances \citep{morris2023weisfeiler}. The primary innovation is the introduction of a cut-based representation, which explicitly models structural partitions to provide a more nuanced predictive signal than standard graph features.

To evaluate the efficacy of these features, we constructed two distinct algorithm selection tasks using instances from the 2023, 2024, and 2025 MiniZinc Challenges.\footnote{\url{https://www.minizinc.org/challenge/}} Experimental results across Support Vector Machines, Random Forests, and Multi-Layer Perceptrons demonstrate that our cut-based features outperform \texttt{fzn2feat} on both tasks when paired with SVMs. Comparisons with RFs and MLPs show smaller differences, with \texttt{fzn2feat} achieving slightly better results in several RF configurations. Our work effectively demonstrates that graph-based structural information can be leveraged to enhance traditional machine learning architectures, ensuring performance gains that are robust across diverse algorithmic structures.

\section{Background}
This section provides the theoretical foundation for this work.

\subsection{Combinatorial Optimisation}
Constraint Programming (CP) \citep{rossi2006handbook} is a paradigm for solving combinatorial optimisation problems (COPs) by defining them via variables, domains, and constraints. CP utilises global constraints (such as \textit{all-different}, \textit{cumulative}, or \textit{circuit}) which capture complex substructures. These constraints use specialised filtering algorithms to prune the search space through propagation, enabling more efficient modelling and solving.

Constraint solvers navigate the solution space using techniques like constraint propagation \citep{bessiere2006constraint}, branch and bound \citep{boyd2007branch}, and conflict-driven clause learning (CDCL) \citep{marques2009conflict}. Given that COPs are often NP-hard \citep{Papadimitriou_book}, parallelisation is frequently employed to speed up the computation. This typically follows two patterns: (i) the job-splitting approach, which parallelises a single solver's strategy, and (ii) the portfolio approach, which runs multiple solvers with different strategies simultaneously \citep{paper_amai}. Comprehensive reviews for parallelising these specific techniques are available in the literature \citep{gent2018review,crainic2006parallel,martins2012overview}. Surveys of algorithm selection are provided by \citet{kerschke2019automated,kotthoff2016algorithm}.

\subsection{The Algorithm Selection Problem}
Algorithm Selection (AS) involves identifying the most suitable algorithm for a specific problem instance to optimise metrics like computational efficiency or accuracy. This process is underpinned by the No Free Lunch theorem \citep{wolpert1997no}, which, under its assumptions, rules out an algorithm that outperforms all others across all objective functions. This motivates selecting algorithms according to the problem context, since solver performance can vary across instances.

Formally, given a problem instance $x \in \mathcal{P}$ drawn from an instance space $\mathcal{P}$ and a portfolio of candidate algorithms $\mathcal{A}$, the objective of an algorithm selector is to determine a mapping $S: \mathcal{P} \to \mathcal{A}$ that optimizes an expected performance metric $m: \mathcal{A} \times \mathcal{P} \to \mathbb{R}$ (such as runtime, penalized average runtime, or solution quality). In practice, machine learning models parameterise this mapping by first transforming the instance $x$ into an informative feature vector $\mathbf{f}(x) \in \mathbb{R}^d$.

\subsection{The WL Algorithm}\label{sec:wl}
The original purpose of the WL algorithm is to check if two graphs are isomorphic \citep{kiefer2020power}. It does so by assigning a colour to each node on a graph and then iteratively refining it by aggregating its colour with the colours of the neighbouring nodes. The algorithm keeps refining the colours until either (a) the two graphs have different colour histograms (meaning they are non-isomorphic) or (b) the refinement does not change the induced partitioning of the graphs (meaning that the graphs are potentially isomorphic).

It is important to note that the WL algorithm is not a complete isomorphism test: two non-isomorphic graphs can remain indistinguishable. For example, equally sized regular graphs of the same degree remain indistinguishable by 1-WL when all nodes initially have the same colour.

It is also possible to generalise the WL algorithm to the $k$-WL algorithm, where the colours are assigned to a $k$-tuple of nodes. In this sense, the standard WL algorithm is a special case of the $k$-WL algorithm where $k=1$. As $k$ grows, the $k$-WL algorithm becomes much more powerful in distinguishing non-isomorphic graphs; however, its computational complexity also grows exponentially in $k$. Thus, for the purpose of this paper, we only focus on the standard 1-WL algorithm.

\section{Related Work}
\label{sec:related-work}
In this section, we review related literature on algorithm selection methods and their feature representations, graph representations of constraint problems, and graph feature extraction techniques.

\subsection{Algorithm Selectors and Handcrafted Features}
Most modern algorithm selectors employ machine learning techniques to predict which algorithm will perform best on an unseen problem instance. In the SAT domain, SATzilla \citep{xu2012satzilla2012} uses Random Forests over portfolios of SAT solvers. In CP, approaches like CPHydra \citep{bridge2012case} and SUNNY \citep{amadini2014sunny} utilise $k$-nearest neighbours to generate instance-specific solver schedules.

A central challenge in algorithm selection is constructing informative feature representations of problem instances. Traditionally, selectors have relied on manually designed, domain-specific features. For instance, SUNNY relies on \texttt{fzn2feat} \citep{Amadini2014-fn}, which extracts static features from a FlatZinc model, such as the counts and types of variables and constraints, domain sizes, and basic metrics derived from the variable-constraint graph (e.g., degree statistics). The \texttt{mzn2feat} 1.2.1 tool, which includes the FlatZinc extractor \texttt{fzn2feat}, provides 95 features.\footnote{\url{https://github.com/CP-Unibo/mzn2feat}} However, such handcrafted features are expensive to design, often fail to generalize across distinct problem domains, and compress complex combinatorial substructures into flat summary statistics.

\subsection{Graph Representations for Constraint Problems}\label{sec:graph-representation}
While representing combinatorial problems as graphs is a conceptually intuitive approach \citep{bockmayr2005constraint}, determining the optimal structural configuration for such graphs remains a non-trivial challenge. Over the years, many different approaches have emerged: a lot of MIP-related tasks employ a bipartite graph representation where a pair of variable/constraint nodes share an edge when the variable has a non-zero coefficient in the constraint \citep{khalil2022mip}. Another approach, often used in CP, is to represent the problem as a tripartite graph, where nodes are constraints, variables and values in the variables' domains. Variable and constraint nodes share edges if the variable participates in the constraint. Furthermore, variable nodes also have edges that connect them to all possible values in their domain \citep{cappart2023combinatorial}. Finally, \citet{boisvert2024towards} decide to decompose constraints into smaller components, and the final graph is akin to the abstract syntax tree of a program.

\subsection{Graph Kernels and GNNs in Combinatorial Optimisation}
Feature extraction transforms graph-structured data into fixed-dimensional vector spaces \citep{graphkernels}. Classic methods rely on graph-theoretical metrics to describe structural properties. Node-level features describe a node's topological importance, including Degree Centrality, Clustering Coefficient \citep{watts1998collective}, and Betweenness or Closeness Centrality \citep{newman2010network}. Edge-level features are primarily used for link prediction and node proximity, such as Common Neighbours, Jaccard Coefficient, Adamic-Adar Index \citep{adamic2003friends}, and Katz Index \citep{katz1953status}.

Graph-level features are often extracted via graph kernels that compute similarity as a scalar product between graphs:
\begin{itemize}
    \item \textbf{Weisfeiler-Lehman (WL) Kernel} \citep{shervashidze2011weisfeiler}: Iteratively relabels nodes based on neighbor signatures (see Section \ref{sec:wl-feature_Extraction} for more details on the extraction process).
    \item \textbf{Graphlet Kernel} \citep{graphkernels}: Counts small, non-isomorphic subgraphs.
    \item \textbf{Random Walk Kernel} \citep{vishwanathan2010graph}: Compares graphs based on matching paths.
\end{itemize}
WL-based features are used in cheminformatics \citep{young2022literature} as well as in neighbouring fields like automated planning \citep{chen2024return}. Graph neural networks have also been applied to computer vision, for example to point cloud classification \citep{simonovsky2017dynamic}.

In recent years, Graph Neural Networks (GNNs) have gained significant popularity for combinatorial optimisation tasks. Crucially, the message-passing mechanism in standard GNNs is fundamentally connected to and bounded in expressive power by the 1-WL test \citep{Xuetal2018,morris2023weisfeiler}. However, while GNNs learn continuous representations via deep architectures, they require substantial training data and incur non-trivial inference latency when querying instances. For algorithm selection, where feature extraction and selector inference must be lightweight to avoid offsetting the performance gains of the selected solver, graph kernels based on Weisfeiler-Leman refinement provide a compelling alternative that preserves theoretical expressiveness while remaining fast to compute.

\section{Graph Conversion}\label{sec:conversion}

In this section we describe the graph conversion we propose in this work. Since the graph conversion is decoupled from the feature extraction methodology, we will separate the two sections and defer the feature extraction methodology discussion to Section \ref{sec:feature_Extraction}. As stated in Section \ref{sec:graph-representation}, many different approaches for graph conversions have been proposed. Our approach is inspired by that of \citet{boisvert2024towards}. We have decided to use Boisvert et al.'s syntax tree-like representation as it allows us to decompose complex constraints into simpler elements that could be reused in different constraints with similar meanings.

Unlike Boisvert et al., we do not include domain values in our representation. We decided to exclude domain values for practical reasons. They argue that a good conversion should be injective (e.g. two distinct instances should never be mapped to the same graph). We do recognise the soundness of this argument, however, our primary objective is to use the features for algorithm selection and, as an example, an instance from the 2025 Minizinc challenge\footnote{\url{https://www.minizinc.org/challenge/2025/results/}} (coming from the cgt problem with instance name cgt\_12\_h27\_r0.05\_s0.5\_3) had an objective function with a domain between -1.932.060 and 1.714.644, resulting in 3.646.705 total possible values just for one domain.
The full instance would also require adding other variables and their domains as well as constraints and operators. A graph with this many nodes would be intractable for our purposes. Furthermore, even assuming the tractability of its conversion, the high number of value nodes would add a lot of noise that could weaken the signal from the remaining nodes. Finally, our graphs are directed instead of undirected. This choice gives shallow local constraint decompositions, making it possible to propagate information along their directed paths with a limited number of aggregation steps.

In more detail, we define five types of nodes:
\begin{itemize}
    \item \textbf{Variable nodes}: representing the variables of the problem. 
    \item \textbf{Solution node}: a single node for each problem that represents the type of problem (Minimisation, Maximisation or Satisfaction).
    \item \textbf{Parameter/Literal nodes}: representing the literal values of the problem (e.g. 1, 2, True, False, etc.).
    \item \textbf{Operator nodes}: representing arithmetic and functional operations between variables, parameters and other operators (e.g. $+$, $\times$, $\div$ etc.) Operator nodes have a sub-category for the type of operator represented.
    \item \textbf{Constraint nodes}: representing relational predicates (e.g. $=$, $\neq$, $<$, $\leq$), logical operations (e.g. $\wedge$, $\vee$, $\neg$), and global constraints (e.g. \texttt{all\_different}, \texttt{table}). Constraint nodes also have a sub-category for the type of constraint the node represents. 
\end{itemize}
All node and sub-node types we have defined for the conversion from FlatZinc can be seen in the Appendix \ref{sec:node-types}.

As previously mentioned, edges between nodes are directed: variables and parameter nodes have no incoming nodes and only point to the operators or constraints they take part in. There is only one exception: for optimisation problems, the variable containing the objective value to optimise has an incoming edge from the Solution node. In case of Satisfaction problems, the solution node is isolated. Operator nodes can have both incoming and outgoing edges. Incoming edges come from variables, parameters and other operators that take part in the operation. Outgoing edges go to other operator nodes or to constraint nodes. Constraint nodes can also have edges to other constraint nodes when they appear as a sub-expression to other constraints. As an example, the expression $(x = 3) \wedge (y \neq 5)$, will have three constraint nodes ($=$, $\wedge$ and $\neq$) with $=$ and $\neq$ having exiting edges to $\wedge$.

Edges are typed as well; in particular, there are three types of edges: 
\begin{itemize}
    \item \textbf{Lead edges}: used to connect variables, parameters and operators to operators and constraints for which the order of the operands does not matter (e.g. sum or equality). Lead Edges are also used for variables, parameters and operators that fall on the left side of an operation or constraint where the order of the operands matters (e.g. subtraction or strictly less than). Lead edges are also used to connect the solution node to the objective variable in optimisation problems.
    \item \textbf{Trailing edges}: Used to connect variables, parameters and operators to operators and constraints for which the order of the operands matters and the operands fall on the right side of the operation.
    \item \textbf{Global edges}: Used to connect variables and parameters to global constraints. This helps to distinguish connections to ordinary constraints from connections to global constraints.
\end{itemize}

\begin{figure}[t]
    \centering
    \begin{subfigure}[c]{.45\columnwidth}
        \begin{lstlisting}
var int x;
var int y;
constraint 3*x <= y;
constraint y < 10;
constraint
  all_different([x, y]);
solve maximise x;\end{lstlisting}
    \end{subfigure}%
    \begin{subfigure}[c]{.45\columnwidth}
        \centering
        \includegraphics[width=.9\textwidth]{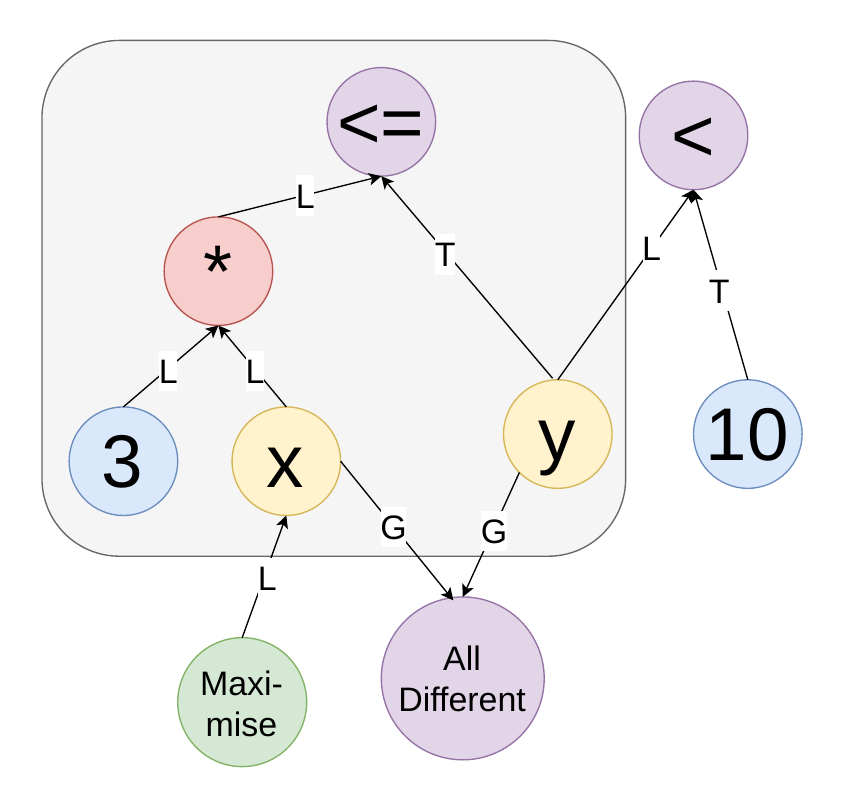}
    \end{subfigure}
    \caption{Conversion of a constraint model (left) to a graph (right). Variables in yellow, Parameters in blue, Operators in red, Constraints in purple, and Solve in green. The area in gray represents the constraint \texttt{3*x <= y}. Lead edges are labelled as L, Trailing as T and Global as G.}
    \label{fig:conversion}
\end{figure}

Figure \ref{fig:conversion} shows an example of how a simple constraint model can be converted using our conversion methodology.

\section{Extracting Features Using The Weisfeiler-Leman Algorithm}\label{sec:feature_Extraction}
Once we have defined how to convert any instance to a graph, we can use the WL algorithm to extract features from the graphs.

\begin{figure}[t]
    \centering
    \includegraphics[width=\columnwidth]{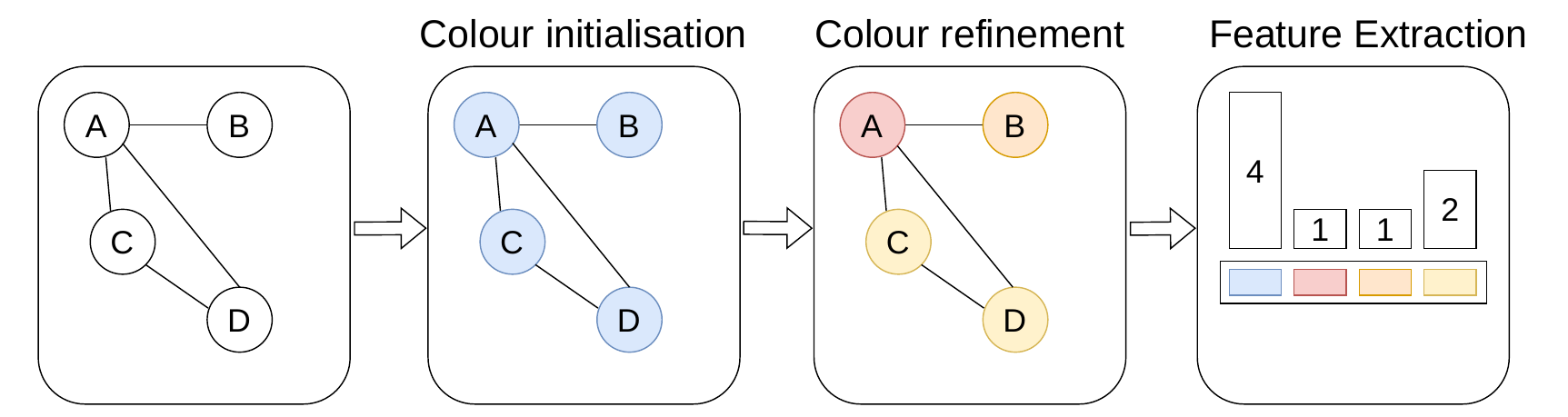}
    \caption{Feature extraction process using the WL algorithm on a simple, undirected graph. The colour refinement is executed only once. At first, all nodes get the same colour, then node A aggregates its colour (blue) with its 3 neighbours' colours (blue), getting, as a result, (blue, blue,blue,blue). This produces a new colour, red. C and D, having only two neighbours, get, as aggregation (blue, blue,blue), resulting in the new colour yellow and, finally, node B, having only one neighbour, aggregates to (blue, blue), getting as new colour orange. In the final steps, we count the number of occurrences of each colour, and we get 4 blues from the original initialisation and 1 red, 1 orange and 2 yellow after the first aggregation step.}
    \label{fig:wl-feature-exec}
\end{figure}

\subsection{WL For Feature Extraction}\label{sec:wl-feature_Extraction}
To use the WL algorithm for feature extraction, we repeat the colour refinement process for a pre-defined number of steps on a single graph, then we extract features by creating a histogram by counting the colors. Then, the final feature vector is created by concatenating the histograms after each iteration. A depiction of the execution of the WL algorithm for feature extraction (with one aggregation step) can be seen in Figure \ref{fig:wl-feature-exec}.

Note that while the standard WL feature extraction process typically concatenates histograms from all iterations, we deliberately utilise only the final histogram. This decision is informed by several theoretical and practical considerations specific to our domain.

First, from a theoretical perspective, the final iteration's labels represent the most refined partition of the graph's nodes. As demonstrated in the analysis of Graph Isomorphism Networks \citep{Xuetal2018}, the final state of a message-passing scheme (given an injective aggregation function) effectively encodes the full subtree structure. In our discrete WL implementation, the final labels serve as a compressed representation that implicitly preserves the structural history of previous iterations.

Furthermore, the specific topology of our data justifies this approach. The extracted graphs are characterised by a shallow diameter and directed edges, which significantly limit the depth of information flow. In such instances, the WL process converges rapidly as nodes quickly incorporate all available predecessors' information. Our empirical analysis (in Appendix \ref{sec:ablation-study}) across two distinct tasks and three different ML models confirmed this: we found no statistically significant difference in performance between concatenated features and the single-histogram approach except for one SVM accuracy comparison, which favoured the single-histogram approach. Consequently, we prioritise the latter.

\subsection{Enhancing Features' Expressiveness}
The simplest WL feature variant starts with identical node colours and does not use edge types. However, in our case, we have both node and edge-level information we can use to better represent the input graph. For this reason, inspired by previous work of \citet{Togninalli2019-qa},  we include two extensions to the base algorithm:
\begin{itemize}
    \item \textbf{Node-information extension}: To add node information into the extraction process, we initialise the node colour with the hash of its type and subtype (e.g. a \texttt{Variable node} and a \texttt{Sum node} will be initialised with different colours). Formally, $c^{(0)}(v) \leftarrow \text{HASH}(\text{type}(v) \ || \ \text{subtype}(v))$ where $||$ denotes string concatenation.
    \item \textbf{Edge-information extension}: To add edge information into the extraction process, we concatenate the edge type to the node colour during the aggregation step (e.g. in the example of Figure \ref{fig:wl-feature-exec}, if node A pointed to node B via a Lead edge, the aggregation of node B would be: (blue, blue-L)). Formally, assuming a directed graph where edges flow towards $v$, its multiset update becomes $M(v) \leftarrow \{\!\!\{c^{(t-1)}(w) \ || \ \text{type}(w, v) \mid (w,v) \in E\}\!\!\}$.
\end{itemize}
Finally, since our graphs are directed, we can filter out outgoing edges and update the color of a node solely based on the incoming edges.

Using these extensions, we can now define four types of features: (i) \texttt{WL}, standard WL features without node or edge information, (ii) \texttt{WLn}, WL features enhanced via node-information extensions, (iii) \texttt{WLe},  WL features enhanced via edge-information extensions, and (iv) \texttt{WLne}, WL features enhanced with both edge and node-information. 

\subsection{The ``Cut''}
In the WL algorithm, the colour of a node is highly dependent on its degree, and two nodes with different degrees, even when sharing the same original colour and the same type of neighbour nodes, will end up with different colours. This can be problematic for nodes representing operators and constraints where the number of connected sub-elements can vary. As an example, let's examine the \texttt{all\_different} global constraint. Using the standard algorithm, \texttt{all\_different([v1, v2, v3])} will get a totally different and uncorrelated colour to \texttt{all\_different([v1, v2])} solely due to its degree. In practice, while preserving some differences between the two constraints, we would still capture the fact that they are similar.

To achieve this, we can virtually ``cut'' the connections between the nodes representing constraints and operators with a variable number of operands in them. We will refer to these nodes as cut nodes. In practice, during the aggregation steps, the cut nodes will not be updated since their incoming connections will be ignored. However, the cut nodes will still contribute to the update of the nodes they are connected to via their outgoing edges. All cut nodes can be seen in Appendix \ref{sec:node-types}. 

After the aggregation steps have been completed, each cut node's colour can be updated by aggregating it with the set (not multiset) comprised of all its connected, original node colours. By using a set rather than a multiset, cut nodes with the same initial colour and the same set of incoming node types will receive the same colour, independent of their in-degree.

The cut shifts the local equivalence criterion from a graded/counting bisimilarity (which tracks exact quantities via multisets) closer to a standard bisimilarity (which tracks only the existence of neighbor types via sets).

Since a cut removes any information regarding the degree of a node, to retain it we add extra features after the colour histogram. Specifically, from each cut node, we can generate node pairs comprising the cut node and its incoming neighbours. Thus, each cut node of in-degree $n$ will generate $n$ pairs. After having generated all the pairs, we count the occurrence of each pair in a graph to generate a histogram of pairs and concatenate it to the histogram of colours.

\begin{figure}[t]
    \centering
    \includegraphics[width=.87\linewidth]{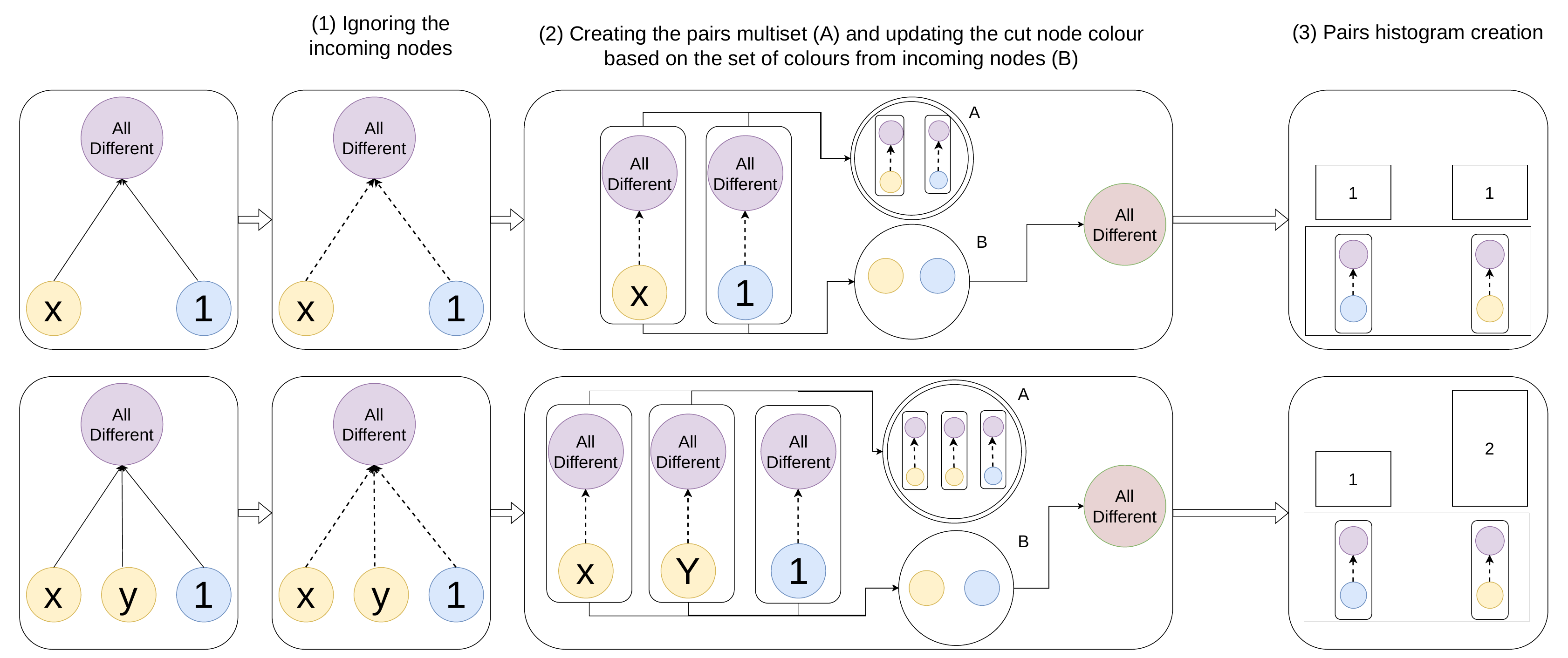}
    \caption{Application of the cut on two occurrences of the \texttt{all\_different} constraint (top and bottom). Constraint nodes are in purple, variable nodes in yellow and parameter nodes in blue. First, cut nodes do not perform the normal WL aggregation steps (1), then we generate all pairs and update the colour of the \texttt{all\_different} node by aggregating it with the set of all types of nodes with incoming edges (2) and then we generate the pairs histogram by counting all the occurrences of the pairs (3). In this example, since both constraints share the same type of incoming nodes, they will be coloured in the same way. However, having a different node degree will result in a different pair histogram.}
    \label{fig:cut-trick}
\end{figure}

The cut can only be applied when node information is present, thus, only to \texttt{WLn} and \texttt{WLne} features. We will call the features where the cut is applied \texttt{WLc-n} if only the cut is applied to \texttt{WLn} features and \texttt{WLc-ne} if the cut is applied to \texttt{WLne}. A full formalisation and line-by-line description of the \texttt{WLc-n} feature extraction procedure is detailed in Algorithm \ref{alg:wl-cut} in Appendix \ref{sec:cut-algorithm}, while a graphical depiction of how the cut is applied can be seen in Figure \ref{fig:cut-trick}.

\subsection{Auxiliary Structural Features}\label{sec:auxiliary_features}
As a consequence of our design choices, especially having directed graphs, variable and parameter nodes have no incoming edges (with the sole exception of the objective variable connected to the Solution node). Nodes of the same source type remain identically coloured within each refinement iteration, so their histogram entries reflect their counts. The objective variable is an exception because it has an incoming edge.

To provide richer structural context beyond raw counts, we augment the histogram representation with two auxiliary features: (i) the average number of constraints and operators in which a variable participates (variable out-degree), and (ii) the average number of operations in which a parameter participates (parameter out-degree).

\subsection{Feature Normalisation and Vocabulary Alignment}\label{sec:normalisation}
Once the raw feature histograms are extracted, two post-processing steps are applied to produce a standardized, model-ready feature representation:

\paragraph{Normalisation.} Raw histogram frequencies scale proportionally with problem instance size. Without normalisation, feature magnitudes reflect instance scale rather than topological structure, hindering cross-instance generalisation. To ensure comparability across diverse problems, we normalise the colour histogram by the total number of nodes $|V|$ and the pairs histogram by $\max(1, |Pairs|)$. This bounds all histogram values within $[0, 1]$.

\paragraph{Vocabulary Alignment.} Across distinct problem instances, the set of generated colours varies, with different graphs sharing only a subset of labels. To construct a consistent vector representation for machine learning models, we assemble a global colour vocabulary across all training graphs. Any colour missing from a particular instance is assigned a frequency of zero. During inference on unseen test instances, any novel colours not observed in the training vocabulary are discarded.

\subsection{Remarks on Graph Dynamics and Depth}\label{sec:feature_characteristics}
Due to the directed nature of our converted graphs, paths from sources to sinks are inherently shallow, especially after applying the cut. For example, in a constraint such as $x + 3 = y$, assuming $x$ and $y$ are not objective variables, information from the source nodes reaches the operator node ($+$) after one iteration and the equality constraint node ($=$) after two. The hash labels can still change in subsequent iterations because each update includes the previous label.

The local constraint decompositions have short directed paths (typically between 1 and 4 edges, with effective propagation depth reduced to 2 under the cut in the example in Appendix \ref{sec:graph-depth}). This local depth should not be confused with the diameter of the entire instance graph. The final set-based update of cut nodes deliberately ignores incoming multiplicities; it therefore differs from standard multiset-based WL refinement.

\section{Experimental Results}

To evaluate the proposed methodology, we implemented the graph conversion algorithm within the FlatZinc modelling ecosystem. We used the 300 instances from the 2023, 2024, and 2025 MiniZinc Challenges to construct two algorithm selection tasks: (i) maximising the Borda count score across three solvers, (ii) maximising prediction accuracy for the same solver set. Solvers were executed with the \texttt{--free-search} flag enabled\footnote{\url{https://docs.minizinc.dev/en/stable/command\_line.html\#cmdoption-f}} on an Intel Xeon Gold 6130 processor with a 20-minute timeout.

Our approach is compared against two baselines: a standard performance baseline (single best solver (SBS) or majority classifier) and \texttt{fzn2feat}.

Given the limited depth of our graph representations (see Section~\ref{sec:feature_characteristics}), we evaluate \texttt{WL} features using one and two aggregation steps. The number of aggregation steps is appended to the features using a dash (\texttt{-}) sign (example: \texttt{WLc-n} features with one aggregation will be \texttt{WLc-n-1}).

For each task, we trained three models: Support Vector Machines (SVM), Random Forests (RF), and Multi-Layer Perceptrons (MLP) with three hidden layers. The selection of SVM, RF, and MLP provides an evaluation framework that tests feature sets against three distinct mathematical paradigms: geometric margin optimisation, ensemble-based logical partitioning, and connectionist functional approximation. All tasks have been evaluated using repeated five-fold nested cross-validation on 10 random seeds for a total of 50 outer-fold evaluations per task per ML model. The training pipeline followed a four-step protocol:
\begin{enumerate}
  \item \textbf{Hyperparameter Optimisation :} We performed a grid search over predefined configurations, evaluating each via mean 5-fold internal cross-validation.
  \item \textbf{Model Selection:} We selected the highest-scoring configuration. Ties were broken by prioritising simpler models to mitigate overfitting:
    \begin{itemize}
      \item For SVM: lower $C$ (stronger regularisation) and $\gamma$ values, favoring the RBF kernel.
      \item For RF: fewer trees and shallower maximum depth.
      \item For MLP: fewer training steps and a smaller $\alpha$ penalty.
    \end{itemize}
  \item \textbf{Dimensionality Reduction:} We evaluated the best configuration on reduced feature sets using Principal Component Analysis (PCA) in the range from $20$ to $\min(\text{n\_features}, \text{n\_samples})$ with a step size of 20.
      \item \textbf{Final Feature Selection:} The feature set (original or PCA-transformed) yielding the highest cross-validation score was selected for final evaluation.
    \end{enumerate}

    The last two steps (Dimensionality Reduction and Final Feature Selection) have been skipped for MLP classifiers due to their ability to handle high-dimensional data.

    All figures group together features extracted with the same methodology but with different aggregation steps, using an alternating shaded background to improve readability.

    \begin{figure}[t]
      \centering
      \noindent
      \includegraphics[width=.7\linewidth]{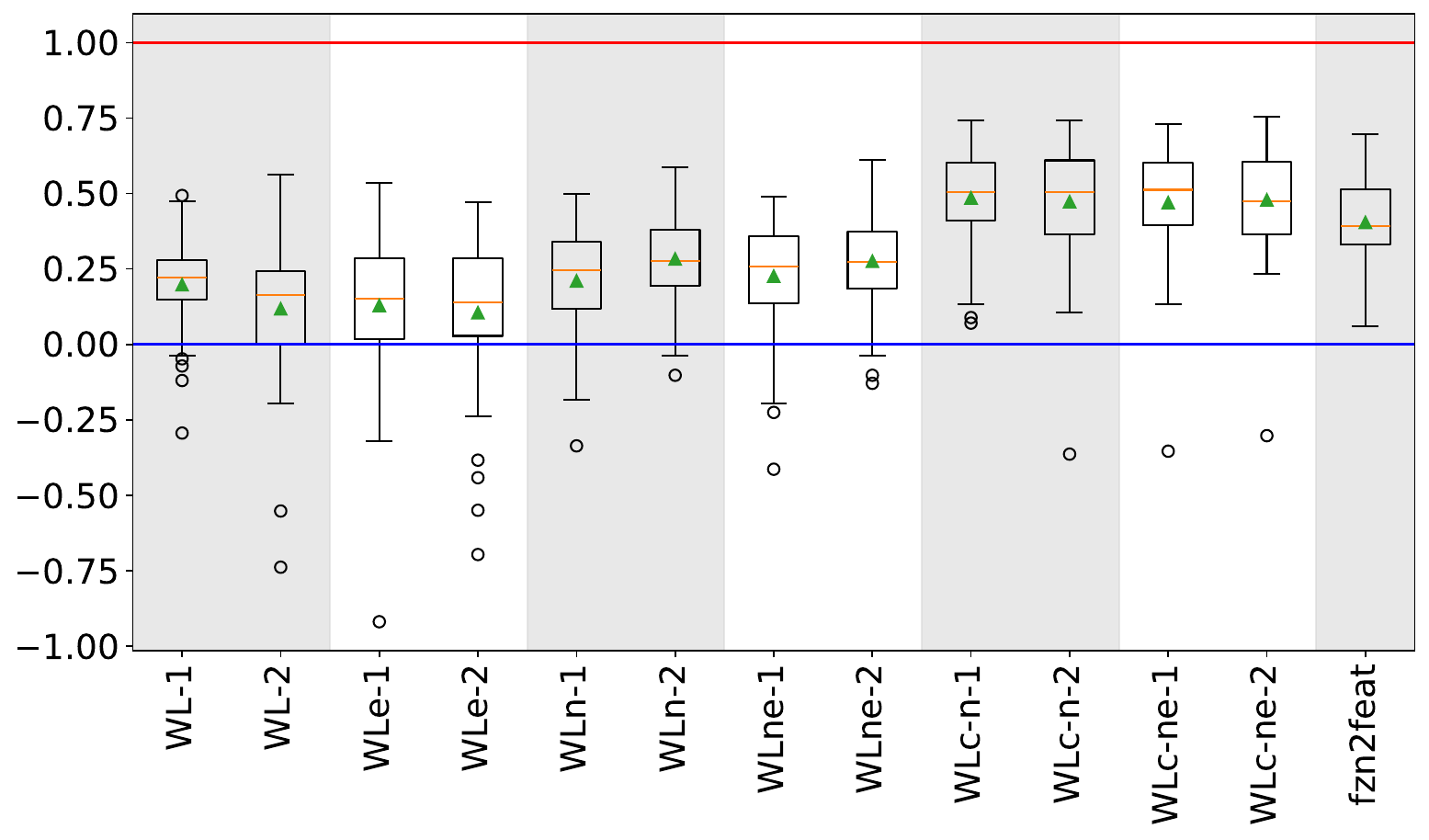}
      \caption{Performance results for SVM models trained on \texttt{WL}-based and \texttt{fzn2feat} features for the Borda count maximization task. Scores (top) are normalized as $S_{norm} = \frac{S - S_{sbs}}{S_{vbs} - S_{sbs}}$, where $S$ is the model score, $sbs$ is the Single Best Solver, and $vbs$ is the Virtual Best Solver. 
      }
      \label{fig:svc-wlc-fzn2feat-scores}
    \end{figure}

    \subsection{Borda Count Score Maximization}\label{sec:borda-cout}
    In this task, we compare three sequential solvers: \texttt{OR-Tools} (CP-SAT) \citep{ortools-cpsat}, \texttt{Chuffed} \citep{chuffed}, and \texttt{CPLEX} \citep{Nickel2022-eb} on the 300 instances. Scores were assigned based on the Borda count strategy used in the MiniZinc Challenge.\footnote{\url{https://www.minizinc.org/challenge/2025/}} We framed this as a classification problem where the target label is the solver achieving the highest score. The results and Borda count scores reflect only the solver execution time, excluding feature extraction costs.

    We excluded one trivially unsatisfiable instance from which features could not be extracted, as well as instances where no solver reached a solution. The final dataset comprised 289 instances. We employed \texttt{StratifiedKFold} from Scikit-learn\footnote{\url{https://scikit-learn.org/stable/modules/generated/sklearn.model\_selection.StratifiedKFold.html}}, stratifying based on the performance gap between the Virtual Best Solver (VBS) and the worst-performing solver in the portfolio.

    Figure \ref{fig:svc-wlc-fzn2feat-scores} illustrates the performance of SVM models trained on \texttt{WL}-based features. While all feature sets generally outperformed the baseline, incorporating node information (\texttt{WLn}) yielded a significant performance increase over standard \texttt{WL} features. Conversely, edge information (\texttt{WLe}) provided no substantial predictive signal. Combining node and edge attributes (\texttt{WLne}) did not further improve results beyond the node-only baseline.

    Notably, features derived from the cut-based representation (\texttt{WLc-n} and \texttt{WLc-ne}) significantly outperformed all other variants. The number of aggregation steps had a negligible impact on final performance across all feature sets.

    MLP classifiers exhibit performance trends similar to those of the SVM on \texttt{WL}-based features. On the other hand, RF ones show an increase when adding the node information, but not a significant jump between \texttt{WLn}/\texttt{WLne} features and the cut ones and these feature sets have similar performance. For readability purposes and since cut features are either on par or better than their non-cut counterpart, we will now focus only on the performance of cut-based features relative to the \texttt{fzn2feat} baseline for these models. We will provide their comprehensive comparison in Appendix \ref{sec:borda-scores}.

    \begin{figure}[t]
      \centering
      \includegraphics[width=.44\linewidth]{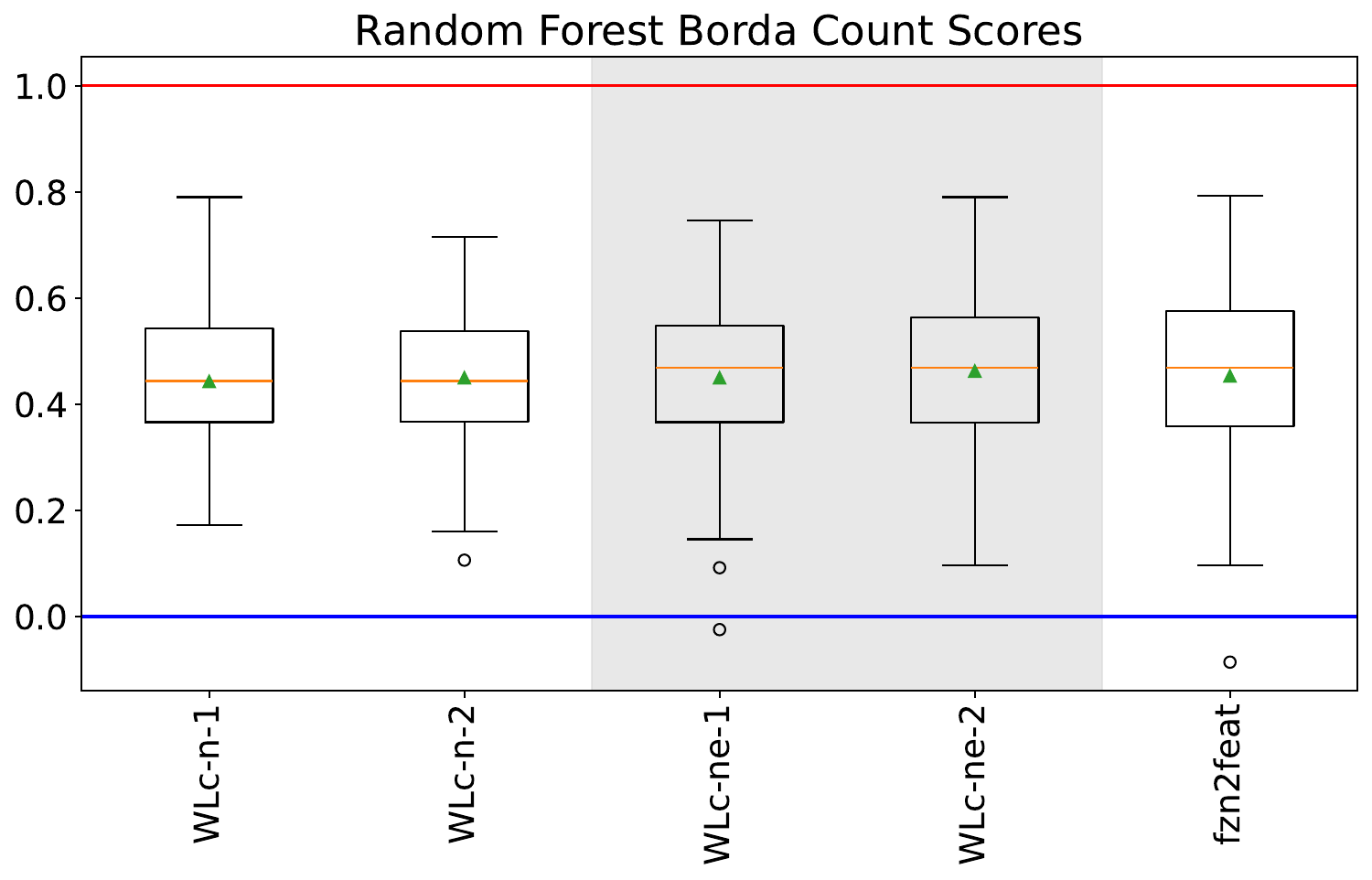}
      \includegraphics[width=.44\linewidth]{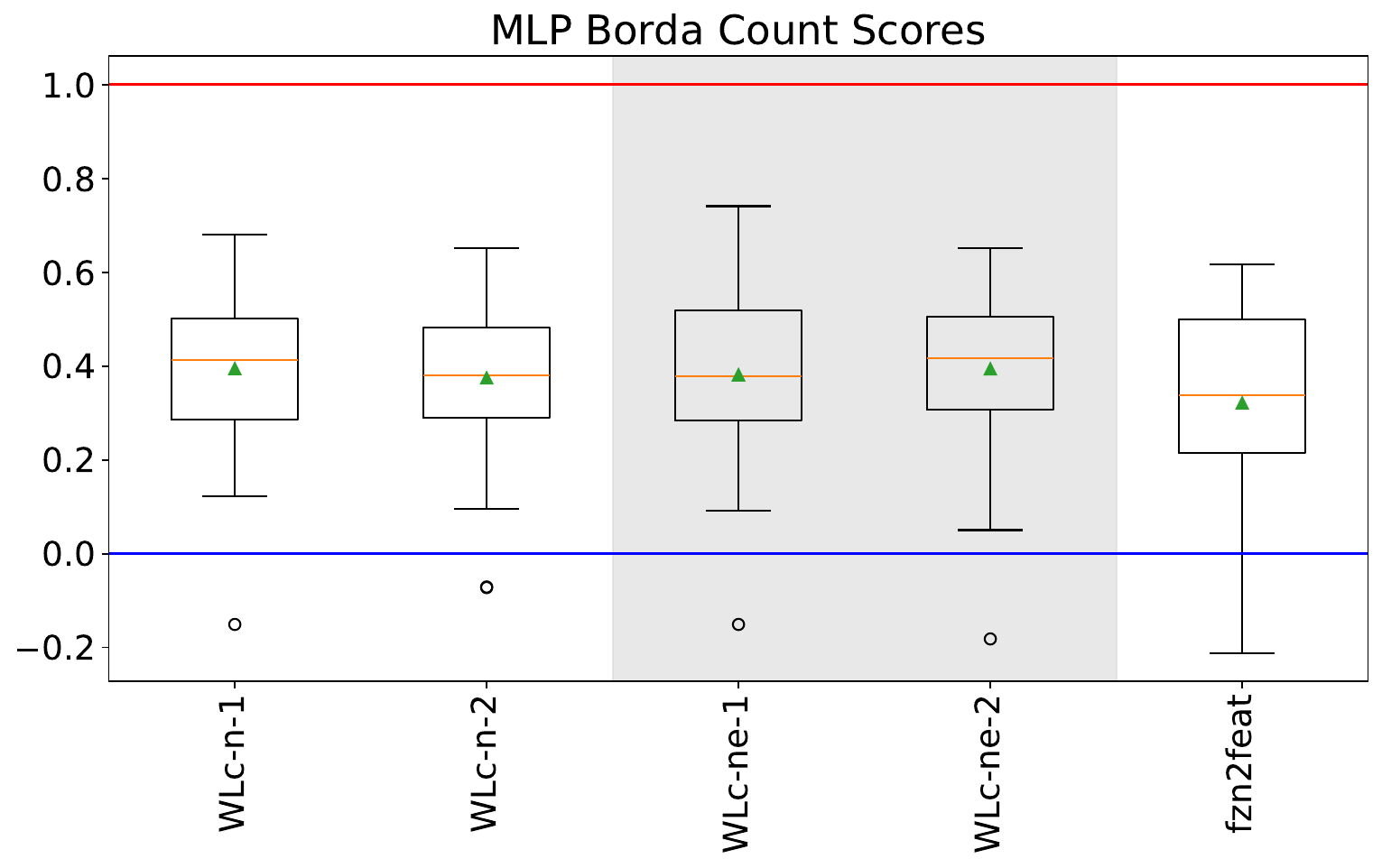}
      \caption{Comparative performance of cut-based features (\texttt{WLc-n/WLc-ne}) against the \texttt{fzn2feat} baseline for the Borda count maximization task. Scores are normalised relative to the Single Best Solver and Virtual Best Solver. Results on the left are obtained by training RF models, while those on the right are obtained by training MLPs.}
      \label{fig:forest-nn-wlc-fzn2feat-scores}
    \end{figure}

    Figure \ref{fig:forest-nn-wlc-fzn2feat-scores} illustrates the comparative performance of RF and MLP models. For the RF, \texttt{fzn2feat} achieves marginally higher scores than the cut-based features, with the exception of \texttt{WLc-ne-2}, which still presents slightly higher mean and median scores.

    \begin{table}
      \centering
      \begin{tabular}{lcc}
        \toprule
        & Mean & Median \\
        \midrule
        \texttt{fzn2feat} & 0.45 & 0.47  \\

        \texttt{Wlc-n-1} & 0.48 & 0.51  \\
        \texttt{Wlc-n-2} & 0.47 & 0.50  \\

        \texttt{Wlc-ne-1} & 0.46 & 0.51 \\
        \texttt{Wlc-ne-2} & 0.48 & 0.48 \\
        \bottomrule
      \end{tabular}
      \caption{Mean and median results of cut features and \texttt{fzn2feat}. Results show the best performing scores obtained using SVM for cut features and RF for \texttt{fzn2feat}.}
      \label{tab:summary-results}
    \end{table}

    Finally, we conducted a cross-model comparison between the best-performing \texttt{fzn2feat} configuration and the best cut-based configuration (SVM) (mean). The results are summarised in Table \ref{tab:summary-results}.

    \subsection{Algorithm Selection with Accuracy Maximisation}
    The setup for this task was the same as for the Borda count maximisation one (in Section \ref{sec:borda-cout}), with the only difference being the objective of the task, which changes from maximising the score of the portfolio to maximising predictive accuracy. In this task, the baseline to beat is the majority classifier, which always predicts the most frequent label.

    The experimental configuration for this task mirrors the Borda count maximisation setup (Section \ref{sec:borda-cout}), with the objective shifted to maximising predictive accuracy. This change of objective was also reflected in the stratification strategy, which is based on the final label instead of the gap.

    \begin{figure}[t]
      \centering
      \includegraphics[width=.7\linewidth]{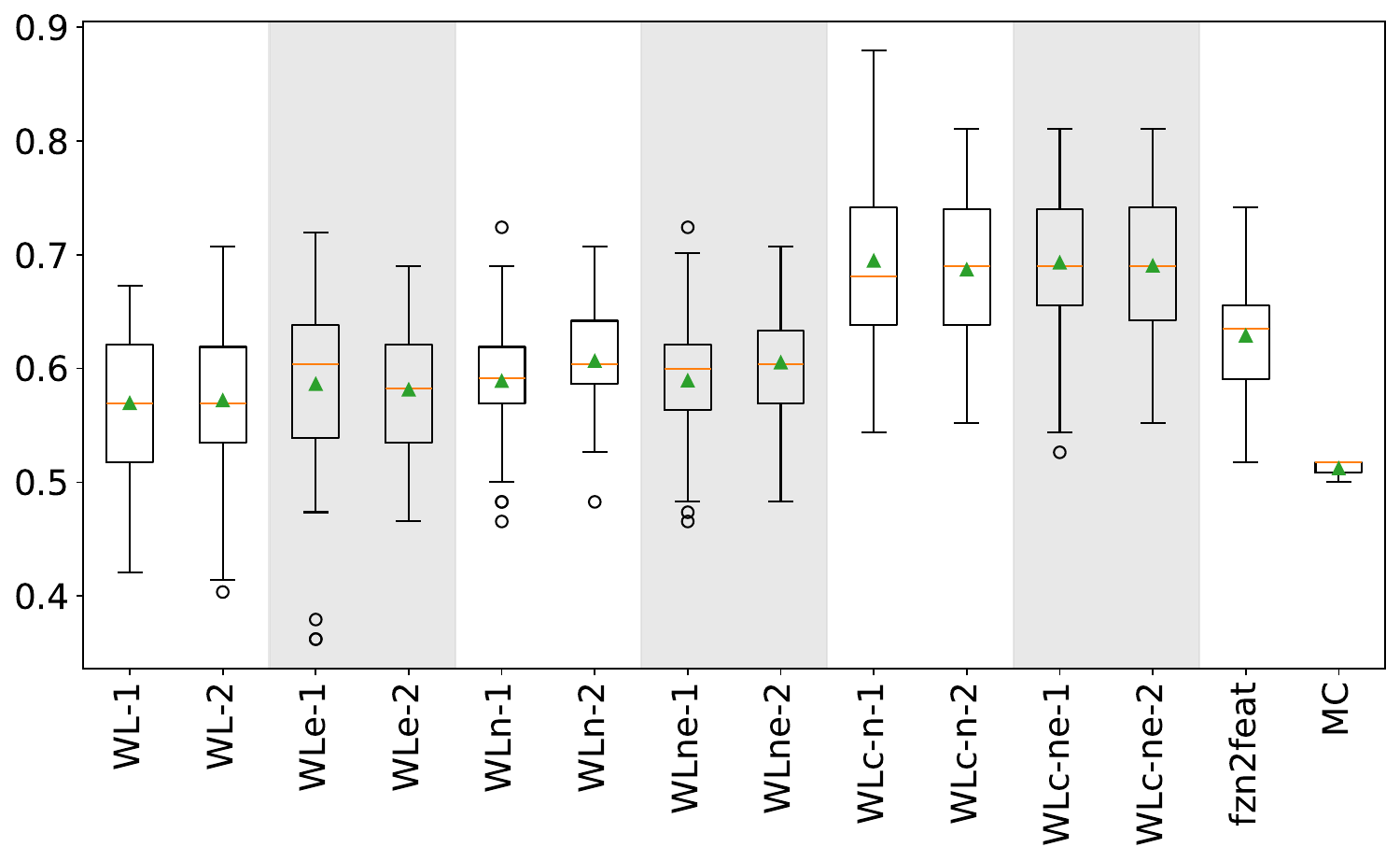}
      \caption{Predictive accuracy for SVM models trained on \texttt{WL}-based and \texttt{fzn2feat} features. 
      }
      \label{fig:svc-accuracy}
    \end{figure}

    Consistent with the observations in Section \ref{sec:borda-cout}, the comparisons of \texttt{WL}-based features for SVM models (Figure \ref{fig:svc-accuracy}) indicate that cut-based features significantly outperform alternative variants. However, for the accuracy maximisation task, the marginal benefit of incorporating node information is less pronounced than in the Borda count task. Regardless, all feature configurations demonstrate a significant improvement over the majority classifier baseline.

    When compared against the \texttt{fzn2feat} baseline (Figure \ref{fig:svc-accuracy}), cut-based features exhibit significantly higher accuracy. 

    As in Section \ref{sec:borda-cout}, RF models show similar performance with cut and non-cut features that include node information.
    However, MLP-based models demonstrate a significant improvement when paired with cut features. 
    We show a comprehensive comparison in Appendix \ref{sec:as-acc}, while here we show only cut-based features comparisons with \texttt{fzn2feat} when paired with RF and MLP models.

    \begin{figure}[t]
      \centering
      \includegraphics[width=.44\linewidth]{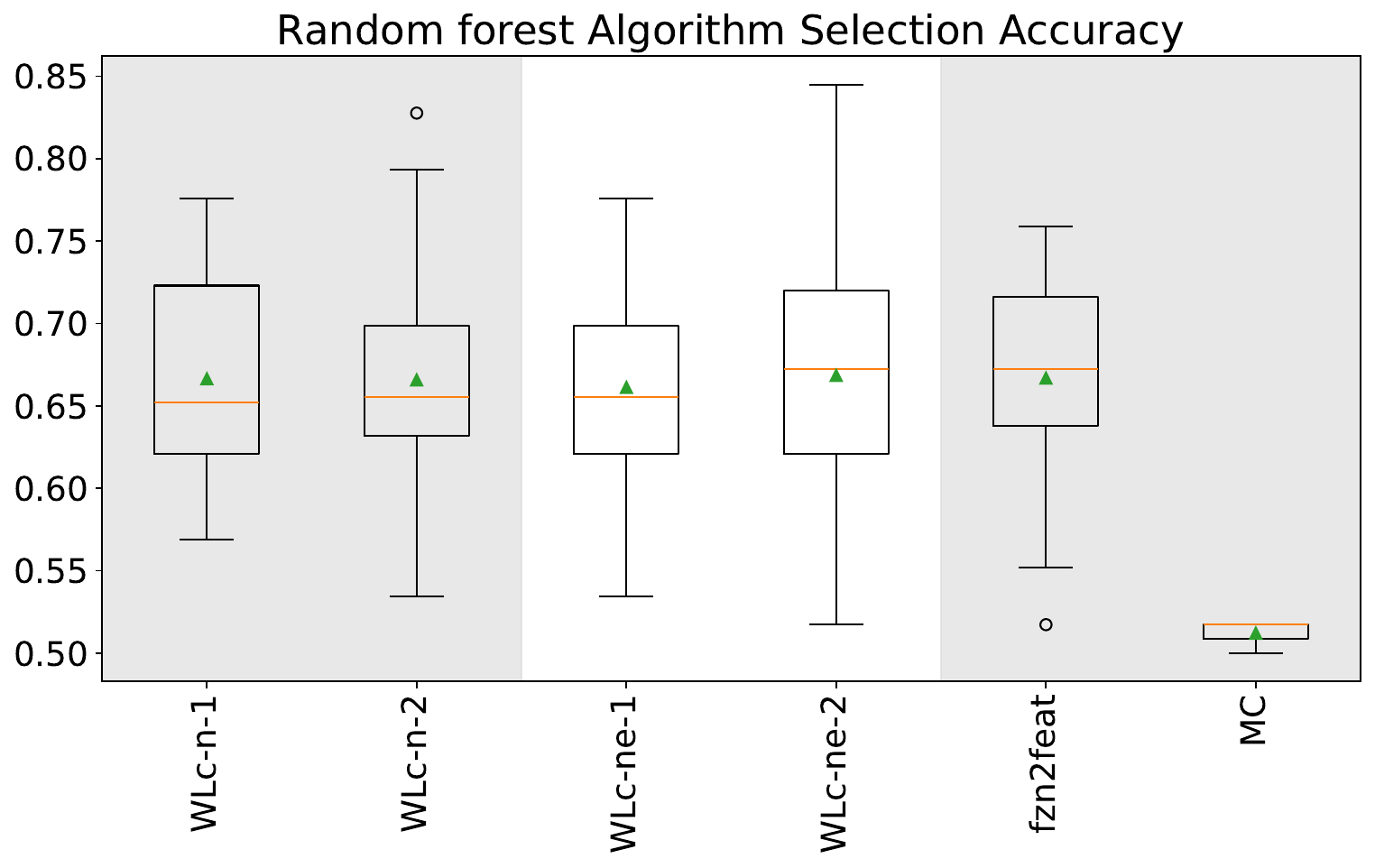}
      \includegraphics[width=.44\linewidth]{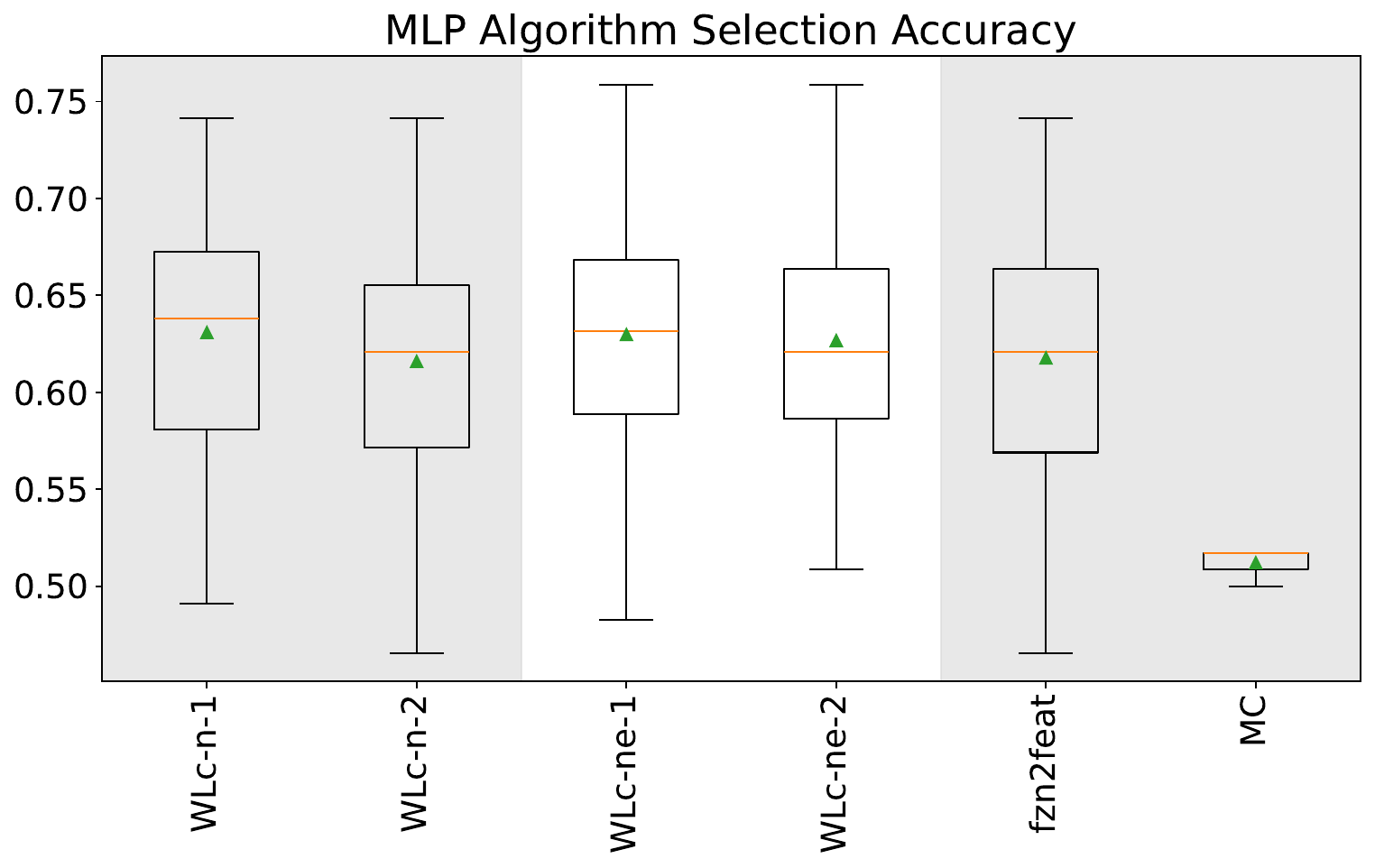}
      \caption{Comparative performance of cut-based features (\texttt{WLc-n/WLc-ne}) against \texttt{fzn2feat} and the majority classifier (MC) baseline. Results on the left are obtained by training RF models, while those on the right are obtained by training MLPs.}
      \label{fig:forest-nn-wlc-fzn2feat-accuracy}
    \end{figure}

    Figure \ref{fig:forest-nn-wlc-fzn2feat-accuracy} shows the performance of RF and MLP-based classifiers trained with cut-based and \texttt{fzn2feat} features. While \texttt{fzn2feat} features perform slightly better on the RF classifier, the differences are minimal.
    Similarly small differences can be seen for the MLP-based classifiers, where all cut-based features perform slightly better than \texttt{fzn2feat}.

    \begin{table}[t]
      \centering
      \begin{tabular}{lcc}
        \toprule
        & Mean & Median \\
        \midrule
        \texttt{fzn2feat} &0.66 & 0.67  \\

        \texttt{Wlc-n-1} & 0.69 & 0.68 \\
        \texttt{Wlc-n-2} & 0.69 & 0.69 \\

        \texttt{Wlc-ne-1} & 0.69 & 0.69 \\
        \texttt{Wlc-ne-2} & 0.69 & 0.69 \\
        \bottomrule
      \end{tabular}
      \caption{Mean and median results of cut features and \texttt{fzn2feat}. Results show the best performing scores obtained using SVM for cut features and RF for \texttt{fzn2feat}.}
      \label{tab:summary-accuracy-results}
    \end{table}


    Table \ref{tab:summary-accuracy-results} pairs the best models for each feature type. Again, cut features obtain better mean and median results than \texttt{fzn2feat}.

    \subsection{Feature Extraction Cost}\label{sec:extraction-cost}
\begin{figure}[t]
      \centering
      \includegraphics[width=.9\linewidth]{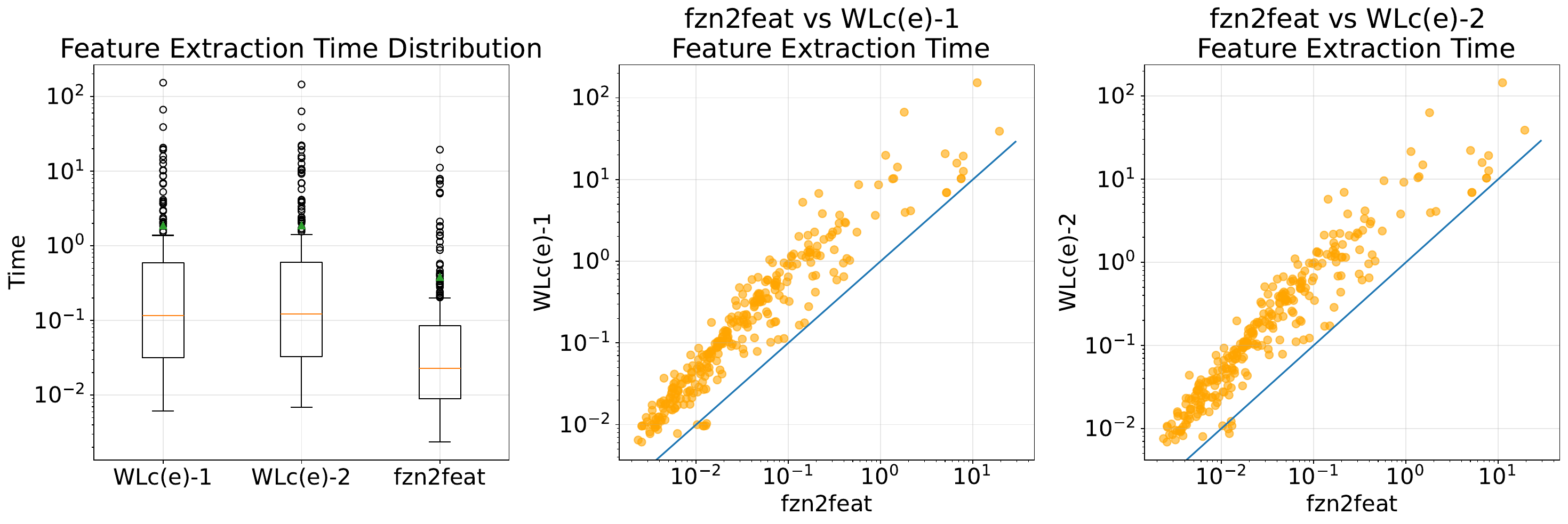}
      \caption{Feature extraction cost comparison between \texttt{WLc-n}/\texttt{WLc-ne} and \texttt{fzn2feat}. The distribution of extraction times (left) is supplemented by pairwise performance comparisons (centre, right). The blue line corresponds to x = y.}
      \label{fig:extraction-time}
    \end{figure}

    The computational overhead for extracting cut-based features compared to \texttt{fzn2feat} is illustrated in Figure \ref{fig:extraction-time}. Because both methodologies necessitate a preliminary compilation from MiniZinc to FlatZinc, we exclude compilation time from the reported extraction costs to isolate the performance of the feature generators themselves.

    While \texttt{fzn2feat} exhibits lower absolute extraction times, the vast majority of instances require less than one second, even when utilising cut-based features. It is important to note that the current implementation of the \texttt{WL} feature extraction pipeline (comprising graph conversion and Weisfeiler-Lehman aggregation) is not fully optimised. Significant performance gains could be realised through algorithmic refinements and the parallelisation of independent graph operations, further reducing the overhead for large-scale instance sets.

\section{Conclusions and Discussion}
This paper introduced a graph-based feature extraction methodology for Constraint Programming instances, with a specific focus on cut-based representations (\texttt{WLc}). By integrating this process directly into the FlatZinc ecosystem, we bridged the gap between structural graph representations and classical machine learning paradigms.

Our findings lead to several key conclusions regarding the efficacy of these features:
(i) Robustness Across Paradigms: The \texttt{WL} and \texttt{WLc} features demonstrate high versatility, providing strong predictive signals across different machine learning models. (ii) Stability and Performance Reliability: The proposed features exhibit high stability across diverse tasks.

Future work will focus on optimising the computational efficiency of the extraction pipeline and exploring the integration of these features into dynamic, online algorithm selection frameworks. Furthermore, we will explore the effect of combining our feature sets with classical feature sets (like \texttt{fzn2feat}) to evaluate if combining different types of signals can improve the predictive performance. We will also focus on generalising our features to different tasks not strictly related to algorithm selection but still relevant in the CP community.

\bibliographystyle{named}
\bibliography{references}

\newpage

\appendix
\section{Node Types for FlatZinc}\label{sec:node-types}
Table \ref{tab:nodes} shows all the node types used to decompose FlatZinc models to graphs. While these nodes have been especially designed around the FlatZinc interface, we believe that they could easily generalise to different languages, with the only exception of global nodes, which are tied to those supported by the solver used for the MiniZinc to FlatZinc conversion (Gecode in our case).
\begin{table*}
  \centering
  \resizebox{\textwidth}{!}{
    \begin{tabular}{|c|c|c|c|c|c|}
      \cline{1-6}
 Name & Type & Subtype & Is Cut & Is Global & Description \\
      \cline{1-6}
 Variable & variable & -- & No & No & Problem variable \\
 Parameter & parameter & -- & No & No & Literal value or parameter \\
 Multiply & operator & mul\_node & No & No & Multiplication operator \\
 Linear Sum & operator & lin\_sum\_node & Yes & No & Weighted sum of variables \\
 Sum & operator & sum\_node & No & No & Sum operator \\
 Equality & constraint & equality\_node & No & No & Equality constraint \\
 Inequality & constraint & inequality\_node & No & No & Inequality constraint \\
 Less than & constraint & le\_node & No & No & Strictly less than constraint \\
 Less or Equal & constraint & leq\_node & No & No & Less than or equal constraint \\
 Imply & constraint & imply\_node & No & No & Logical implication \\
 If and only if & constraint & iff\_node & No & No & Logical equivalence \\
 Not & constraint & not\_node & No & No & Logical negation \\
 Or & constraint & or\_node & No & No & Logical disjunction \\
 And & constraint & and\_node & No & No & Logical conjunction \\
 Xor & constraint & xor\_node & No & No & Logical exclusive or \\
 Index & operator & index\_node & No & No & Array index operator \\
 Abs & operator & abs\_node & No & No & Absolute value operator \\
 Division & operator & division\_node & No & No & Integer division operator \\
 Max & operator & max\_node & No & No & Maximum operator \\
 Min & operator & min\_node & No & No & Minimum operator \\
 Modulo & operator & modulo\_node & No & No & Modulo operator \\
 Pow & operator & pow\_node & No & No & Power operator \\
 In & constraint & in\_node & No & No & Set membership constraint \\
 Card & operator & card\_node & No & No & Set cardinality operator \\
 Difference & operator & diff\_node & No & No & Set difference operator \\
 Intersect & operator & intersect\_node & No & No & Set intersection operator \\
 Subset & constraint & subset\_node & No & No & Set subset constraint \\
 Symmetric difference & operator & symdiff\_node & No & No & Set symmetric difference operator \\
 Union & operator & union\_node & No & No & Set union operator \\
 Maximise & solve & maximise\_node & No & No & Maximisation objective \\
 Minimise & solve & minimise\_node & No & No & Minimisation objective \\
 Array & operator & array\_node & No & No & Array container \\
 Cumulative & constraint & cumulative\_node & Yes & Yes & Global cumulative constraint \\
 Integer Element & operator & int\_element\_node & Yes & Yes & Global element constraint \\
 Linear equality iff & constraint & int\_lin\_eq\_impt\_node & Yes & No & Reified linear equality \\
 Array max & constraint & array\_int\_maximum\_node & Yes & Yes & Global array maximum constraint \\
 Schedule unary & constraint & schedule\_unary\_node & Yes & Yes & Global unary scheduling constraint \\
 Linear less than iff & constraint & int\_le\_imp\_node & Yes & No & Reified linear less than \\
 Maximum argument offset & constraint & maximum\_arg\_int\_offset\_node & Yes & Yes & Global maximum argument constraint \\
 Circuit & constraint & circuit\_node & Yes & Yes & Global circuit constraint \\
 Count equality & constraint & count\_eq\_node & Yes & Yes & Global count equality constraint \\
 Count equality iff & constraint & count\_eq\_reif\_node & Yes & Yes & Reified count equality constraint \\
 Set in iff & constraint & set\_in\_imp\_node & Yes & Yes & Reified set membership constraint \\
 Global cardinality & constraint & global\_cardinality\_node & Yes & Yes & Global cardinality constraint \\
 Global cardinality low up & constraint & global\_cardinality\_low\_up\_node & Yes & Yes & Global cardinality constraint with bounds \\
 Global cardinality low up closed & constraint & global\_cardinality\_low\_up\_closed\_node & Yes & Yes & Global cardinality constraint with closed bounds \\
 Xor iff & constraint & bool\_xor\_imp\_node & Yes & No & Reified exclusive or constraint \\
 No overlap & constraint & nooverlap\_node & Yes & Yes & Global no overlap constraint \\
 Regular & constraint & regular\_node & Yes & Yes & Global regular constraint \\
 All different & constraint & all\_different\_node & Yes & Yes & Global all different constraint \\
 Equality iff & constraint & eq\_imp\_node & Yes & Yes & Reified equality constraint \\
 All equal & constraint & all\_equal\_node & Yes & Yes & Global all equal constraint \\
 Boolean Element & operator & bool\_element\_node & Yes & No & Global element constraint (boolean) \\
 Array minimum & constraint & array\_int\_minimum\_node & Yes & No & Global array minimum constraint \\
 Bool clause iff & constraint & bool\_clause\_reif\_node & Yes & No & Reified boolean clause constraint \\
 Element 2d & operator & int\_element2d\_node & Yes & Yes & Global 2D element constraint \\
 Bin packing load & constraint & bin\_packing\_load\_node & Yes & Yes & Global bin packing load constraint \\
 Table & constraint & table\_int\_node & Yes & Yes & Global table constraint \\
 Precede & constraint & precede\_node & Yes & Yes & Global precede constraint \\
 Array less or equal & constraint & array\_int\_lq\_node & Yes & No & Global array less or equal constraint \\
 Linear less than iff & constraint & int\_lin\_le\_imp\_node & Yes & No & Reified linear less than \\
 Linear not equal iff & constraint & int\_lin\_ne\_imp\_node & Yes & No & Reified linear not equal \\
 Increasing integer & constraint & increasing\_int\_node (int) & Yes & Yes & Global increasing integer constraint \\
 Inverse & constraint & inverse\_offsets\_node & Yes & Yes & Global inverse constraint \\
 N value & constraint & nvalue\_node & Yes & Yes & Global n-value constraint \\
 Not equal iff & constraint & int\_ne\_imp\_node & Yes & No & Reified not equal constraint \\
 Increasing bool & constraint & increasing\_bool\_node & Yes & Yes & Global increasing boolean constraint \\
 Member integer & constraint & member\_int\_node & Yes & Yes & Global member integer constraint \\
 Table iff & constraint & table\_int\_imp\_node & Yes & Yes & Reified global table constraint \\
 At least & constraint & at\_least\_node & Yes & Yes & Global at least constraint \\
 At most & constraint & at\_most\_node & Yes & Yes & Global at most constraint \\
 Integer pow & constraint & int\_pow\_node & Yes & Yes & Global integer power constraint \\
 Global cardinality closed & constraint & global\_cardinality\_closed\_node & Yes & Yes & Global cardinality constraint closed \\
      \cline{1-6}
    \end{tabular}
 }
  \caption{Node types used in our graphs converted from FlatZinc.}
  \label{tab:nodes}
\end{table*}

\section{The Cut Algorithm}\label{sec:cut-algorithm}
Algorithm \ref{alg:wl-cut} outlines the feature extraction procedure for \texttt{WLc-n}. The edge-enhanced variant \texttt{WLc-ne} operates identically, except that edge types are included alongside neighbour colours during aggregation (Line \ref{algline:agg}).

\begin{algorithm}[t]
\small
\caption{\texttt{WLc-n} Algorithm for Feature Extraction.}
\label{alg:wl-cut}
\begin{algorithmic}[1]
\Require Directed graph $G = (V, E)$, iterations $T \geq 1$, cut nodes $V_{cut} \subseteq V$.
\Ensure Feature histograms of colours and pairs.
\State \textbf{Initialisation:} $c^{(0)}(v) \leftarrow \text{HASH}(\text{type}(v) \,||\, \text{subtype}(v))$ for all $v \in V$ \label{algline:init}
\State $t \leftarrow 0$
\Repeat
    \State $t \leftarrow t + 1$
    \For{each node $v \in V$}
        \If{$v \notin V_{cut}$} \label{algline:ifcut}
            \State $M(v) \leftarrow \{\!\!\{c^{(t-1)}(w) \mid (w,v) \in E\}\!\!\}$ \label{algline:agg}
            \State $c^{(t)}(v) \leftarrow \text{HASH}(c^{(t-1)}(v), \text{sort}(M(v)))$ \label{algline:hash}
        \Else
            \State $c^{(t)}(v) \leftarrow c^{(t-1)}(v)$
        \EndIf
    \EndFor
\Until{$t = T$}
\State $Pairs \leftarrow \{\!\!\{\}\!\!\}$ \Comment{Multiset for pair occurrences} \label{algline:pairsinit}
\For{each node $v \in V_{cut}$} \label{algline:forcut}
    \State $S(v) \leftarrow \{ c^{(0)}(w) \mid (w,v) \in E \}$ \Comment{Set of initial incoming neighbour types} \label{algline:set}
    \State $c^{(final)}(v) \leftarrow \text{HASH}(c^{(T)}(v), \text{sort}(S(v)))$ \label{algline:cfinalcut}
    \For{each node $w$ where $(w,v) \in E$} \label{algline:forinw}
        \State $Pairs \leftarrow Pairs \cup \{\!\!\{ (c^{(0)}(w), c^{(0)}(v)) \}\!\!\}$ \label{algline:pairsadd}
    \EndFor
\EndFor
\State $c^{(final)}(v) \leftarrow c^{(T)}(v)$ for all $v \in V \setminus V_{cut}$ \label{algline:cfinalnoncut}
\State \Return $\text{Histogram}(c^{(final)}) \mathbin{||} \text{Histogram}(Pairs)$ \label{algline:return}
\end{algorithmic}
\end{algorithm}

The execution of Algorithm \ref{alg:wl-cut} operates as follows:
\begin{itemize}
    \item \textbf{Line \ref{algline:init} (Initialisation):} Each node $v \in V$ receives an initial colour $c^{(0)}(v)$ based on its type and subtype (e.g., variable, parameter, or operator/constraint type).
    \item \textbf{Lines 3--14 (Iterative Refinement):} Standard Weisfeiler-Leman refinement runs for $T$ steps over non-cut nodes ($v \notin V_{cut}$), updating their colour by hashing the sorted multiset $M(v)$ of incoming neighbour colours (Lines \ref{algline:agg}--\ref{algline:hash}). Cut nodes skip this aggregation (Line \ref{algline:ifcut}), preventing degree-dependent updates, but continue to propagate their colours downstream.
    \item \textbf{Lines \ref{algline:forcut}--\ref{algline:pairsadd} (Cut Node Aggregation \& Pair Counting):} After refinement, cut nodes aggregate the \emph{set} (rather than multiset) $S(v)$ of their incoming neighbour types (Line \ref{algline:set}), updating their final colour in Line \ref{algline:cfinalcut}. This ensures cut nodes of the same type with the same set of incoming node types receive the same colour regardless of arity. Lines \ref{algline:forinw}--\ref{algline:pairsadd} recover the discarded degree information by collecting directed $(neighbour, cut)$ pairs into $Pairs$.
    \item \textbf{Lines \ref{algline:cfinalnoncut}--\ref{algline:return} (Feature Extraction):} Non-cut nodes retain their refined colours $c^{(T)}(v)$, and the final feature vector is formed by concatenating the histograms of node colours and neighbour pairs.
\end{itemize}

\section{On The Shallowness of the Graphs}\label{sec:graph-depth}
\begin{figure}[t]
  \centering
  \includegraphics[width=.5\textwidth]{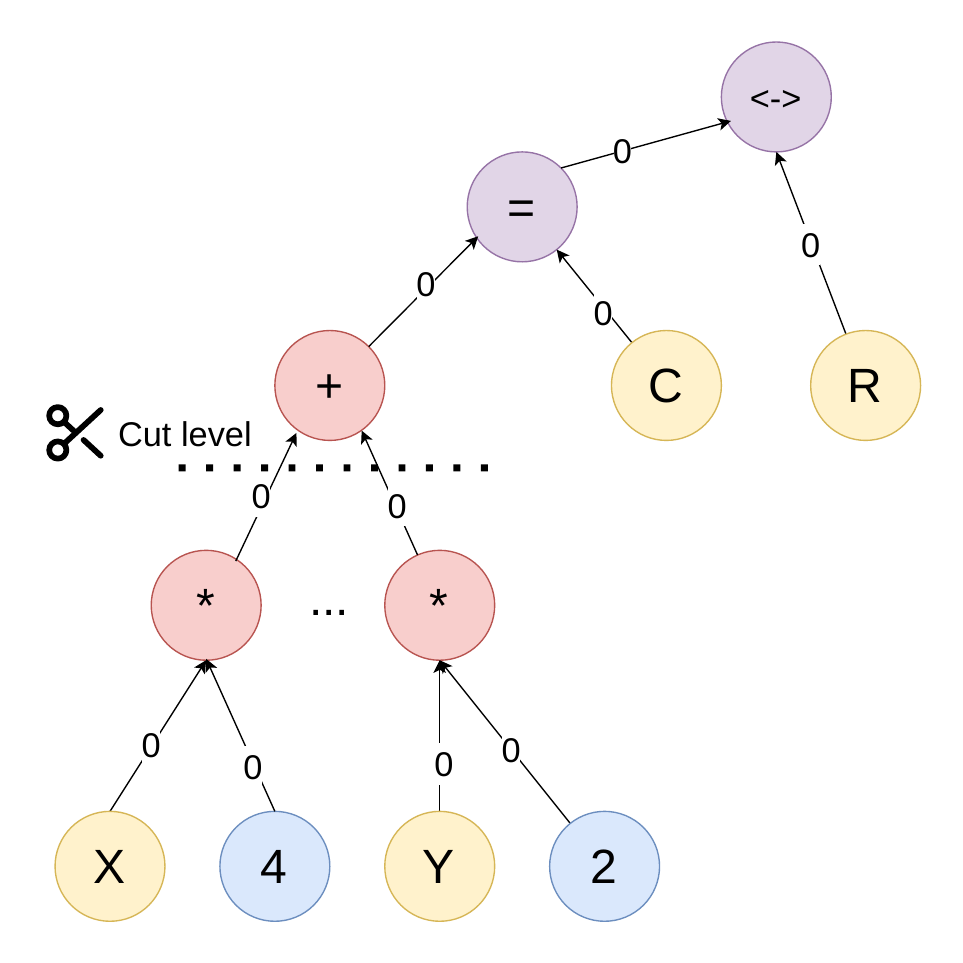}
  \caption{Representation of the deepest constraint currently present in the graph conversion. The constraint corresponds to $\sum_i var_i \times par_i = C \leftrightarrow R$. Variables are in yellow, parameters in blue, operators in red and constraints in purple.}
  \label{fig:long-graph}
\end{figure}

Figure \ref{fig:long-graph} shows the constraint that produces the deepest path currently implemented. A family of similar constraints is present, and they usually differ in the type of constraint that replaces the equality (e.g. inequality, less than or equal to, etc.) This local decomposition has a longest directed path of 4 edges, with effective propagation depth reduced to 2 when the cut is applied. Most supported constraints are, instead, quite simple, with most of them having directed paths of at most one or two edges. For the full list of supported constraints, we refer to the FlatZinc specification.\footnote{\url{https://docs.minizinc.dev/en/stable/lib-flatzinc.html}}

\section{All Borda Count Results}\label{sec:borda-scores}
\begin{figure}
  \centering
  \includegraphics[width=.7\linewidth]{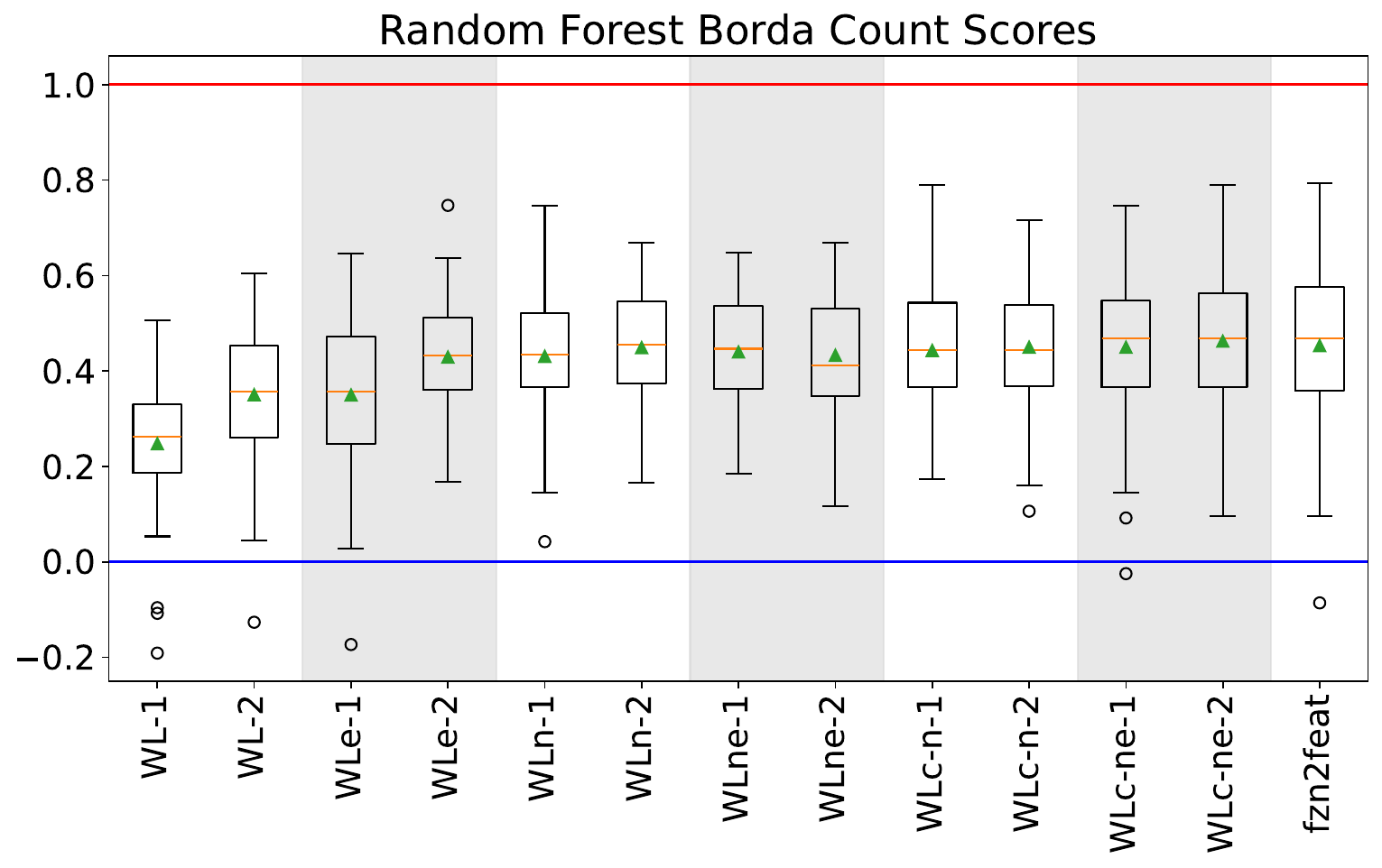}
  \caption{Results of all feature sets trained with the RF classifier on the Borda count maximisation task.}
  \label{fig:RF-bordacount-all}
\end{figure}

\begin{figure}
  \centering
  \includegraphics[width=.7\linewidth]{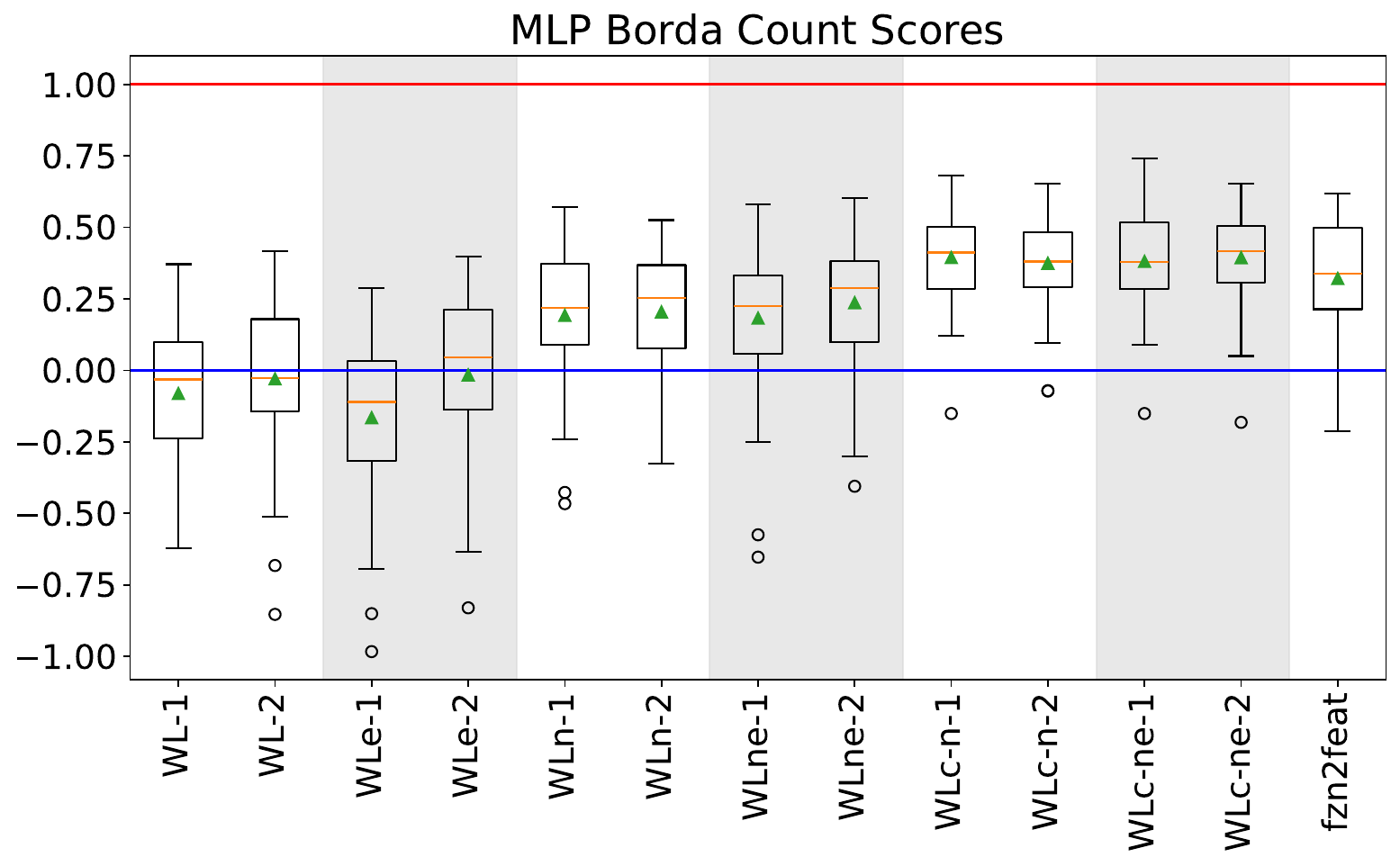}
  \caption{Results of all feature sets trained with the MLP classifier on the Borda count maximisation task.}
  \label{fig:nn-bordacount-all}
\end{figure}

Figures \ref{fig:RF-bordacount-all} and \ref{fig:nn-bordacount-all} show the results on the Borda count tasks for (respectively) RF and MLP classifiers trained with all Feature sets. RF results show similar results for all \texttt{WL} based features that include node information, as well as \texttt{fzn2feat} features. On the other hand, MLP results show a trend similar to SVM, where cut features significantly outperform the other \texttt{WL}-based features, as well as showing a higher gap with respect to \texttt{fzn2feat}.

\section{All Algorithm Selection Accuracy Results}\label{sec:as-acc}
\begin{figure}
  \centering
  \includegraphics[width=.7\linewidth]{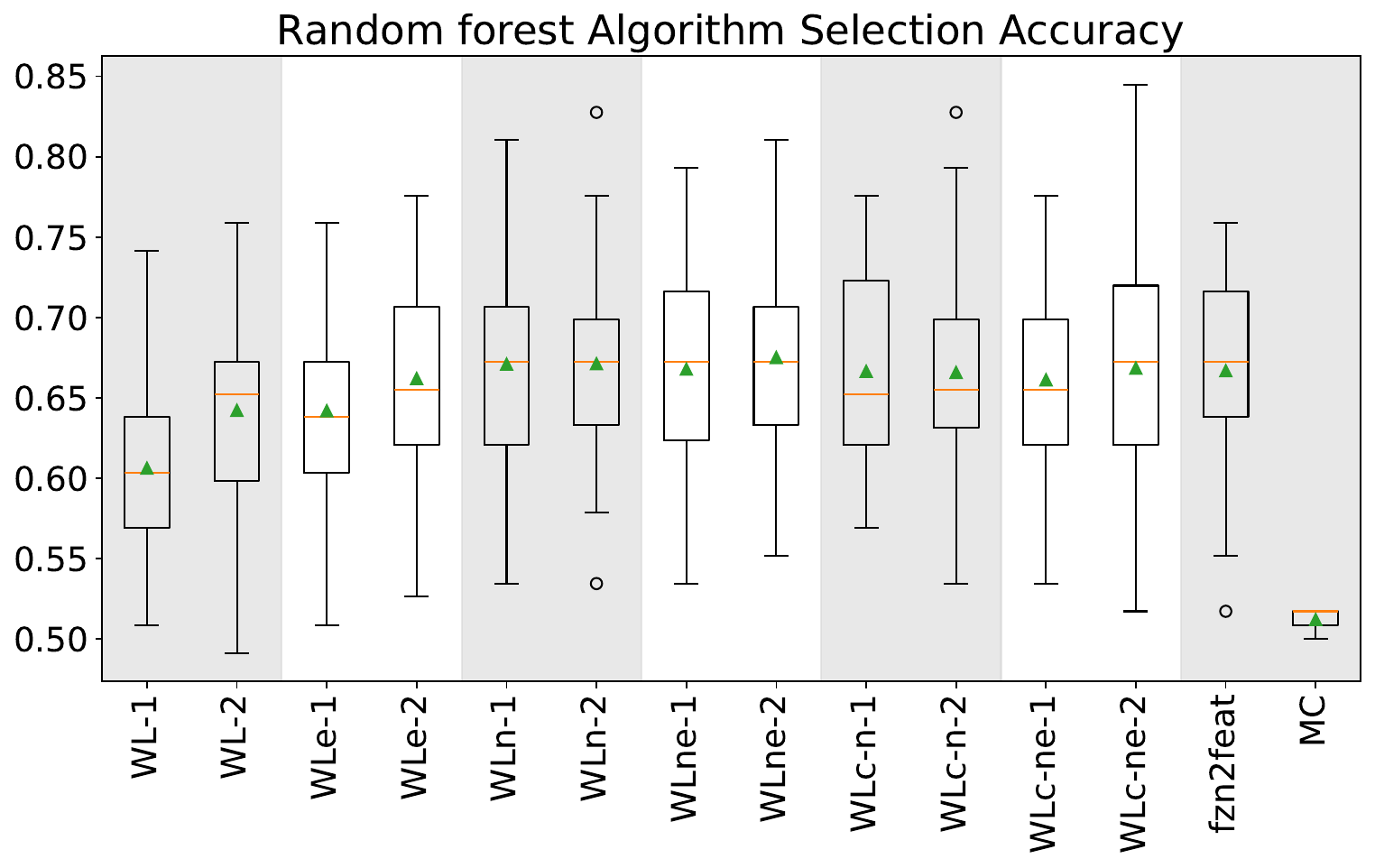}
  \caption{Results of all feature sets trained with the RF classifier on the Algorithm Selection accuracy maximisation task.}
  \label{fig:RF-acc-all}
\end{figure}

\begin{figure}
  \centering
  \includegraphics[width=.7\linewidth]{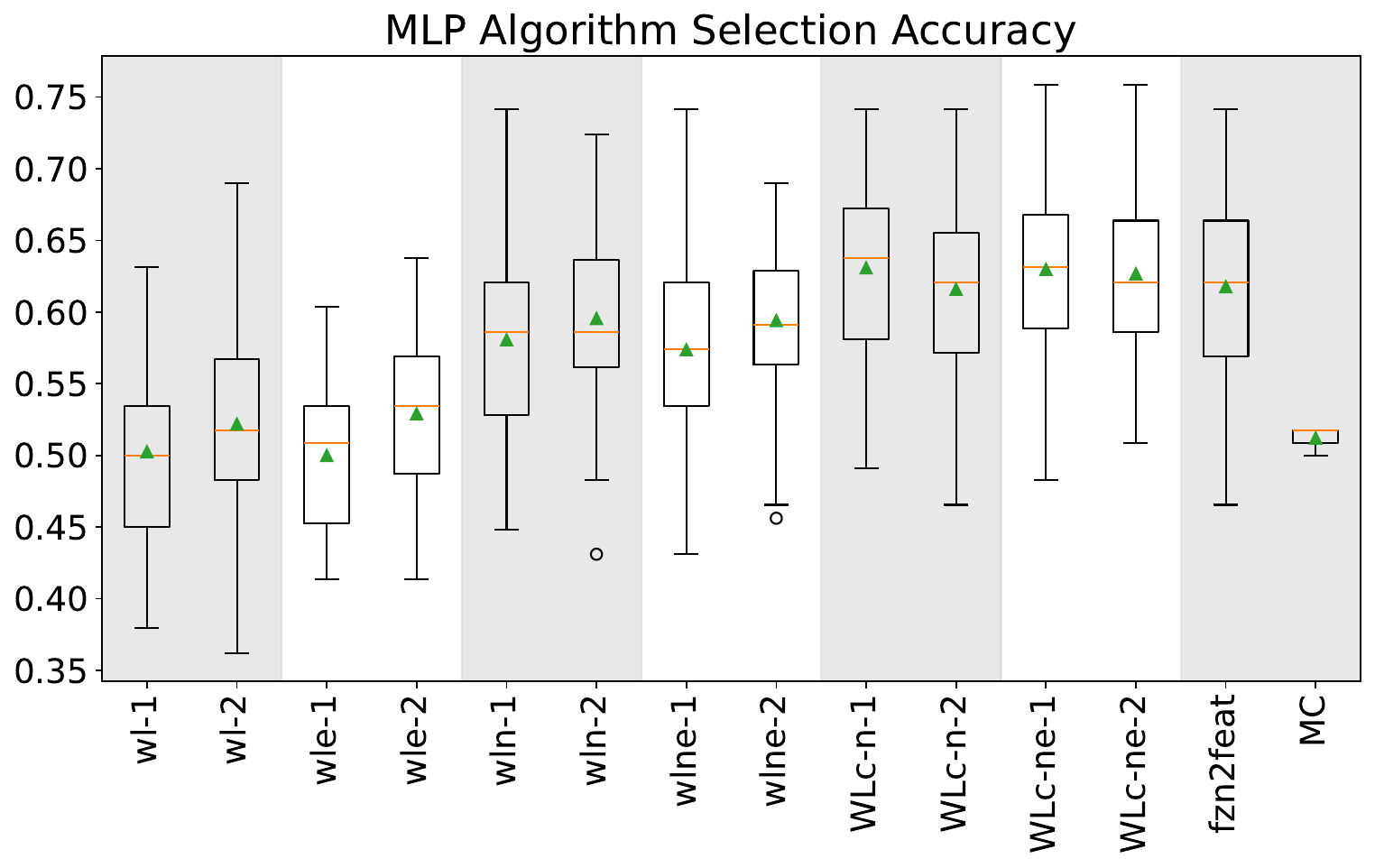}
  \caption{Results of all feature sets trained with the MLP classifier on the Algorithm Selection accuracy maximisation task.}
  \label{fig:nn-acc-all}
\end{figure}

Figures \ref{fig:RF-acc-all} and \ref{fig:nn-acc-all} show the results on the Algorithm selection accuracy tasks for (respectively) RF and MLP classifiers trained with all Feature sets. Similarly to the Borda count counterpart, RF results show similar results for all \texttt{WL} based features that include node information as well as \texttt{fzn2feat} features, while MLP results show a trend similar to SVM where cut features significantly outperform the other \texttt{WL}-based features as well as showing a higher gap with respect to \texttt{fzn2feat} (although less pronounced than in the Borda count task).

\section{Ablation Study on the Importance of Concatenating All Aggregation Levels}\label{sec:ablation-study}
To assess whether only using the last aggregation level of the \texttt{WL} resulted in a loss of performance, we conducted an ablation study on the best-performing features: \texttt{WLc-n-1/2}, \texttt{WLc-ne-1/2}, \texttt{WLn-1/2}, and \texttt{WLne-1/2} across all different tasks and ML models. Results can be found in Tables \ref{tab:all-levels-borda} (Borda score), \ref{tab:all-levels-accuracy} (algorithm selection accuracy), and \ref{tab:all-levels-par} (parallel prediction). The tables show the mean and median results of the ML models trained on only the last level or on all levels of aggregation, as well as the p-value computed with the Wilcoxon test. In all cases but one, the results are not statistically significant. The only case in which they are (\texttt{WLc-n-1} in SVM algorithm selection accuracy prediction), it is in favour of the features with only the last aggregation level.

\begin{table}[t]
  \centering
  \begin{subtable}{.49\linewidth}
    \centering
    \caption{Results of SVM models.}
    \resizebox{\linewidth}{!}{
      \begin{tabular}{llllll}
        \toprule
        & Last mean & All-levels mean & Last median & All-levels median & p value \\
        \midrule
        \texttt{WLn-1} & 0.210 & 0.196 & 0.245 & 0.237 & 0.632 \\
        \texttt{WLn-2} & 0.283 & 0.243 & 0.276 & 0.266 & 0.357 \\
        \texttt{WLne-1} & 0.226 & 0.212 & 0.258 & 0.245 & 0.561 \\
        \texttt{WLne-2} & 0.276 & 0.236 & 0.275 & 0.266 & 0.201 \\
        \texttt{WLc-n-1} & 0.485 & 0.441 & 0.506 & 0.473 & 0.399 \\
        \texttt{WLc-n-2} & 0.472 & 0.432 & 0.504 & 0.456 & 0.198 \\
        \texttt{WLc-ne-1} & 0.470 & 0.440 & 0.513 & 0.487 & 0.681 \\
        \texttt{WLc-ne-2} & 0.479 & 0.432 & 0.475 & 0.460 & 0.318 \\
        \bottomrule
      \end{tabular}
 }
  \end{subtable}

  \vspace{1em} 

  \begin{subtable}{.49\linewidth}
    \centering
    \caption{Results of RF models.}
    \resizebox{\linewidth}{!}{
      \begin{tabular}{llllll}
        \toprule
        & Last mean & All-levels mean & Last median & All-levels median & p value \\
        \midrule
        \texttt{WLn-1} & 0.431 & 0.464 & 0.434 & 0.486 & 0.318 \\
        \texttt{WLn-2} & 0.449 & 0.464 & 0.454 & 0.465 & 0.745 \\
        \texttt{WLne-1} & 0.440 & 0.452 & 0.447 & 0.455 & 0.509 \\
        \texttt{WLne-2} & 0.433 & 0.475 & 0.412 & 0.488 & 0.114 \\
        \texttt{WLc-n-1} & 0.443 & 0.460 & 0.444 & 0.465 & 0.608 \\
        \texttt{WLc-n-2} & 0.450 & 0.449 & 0.444 & 0.459 & 0.863 \\
        \texttt{WLc-ne-1} & 0.450 & 0.456 & 0.469 & 0.484 & 0.774 \\
        \texttt{WLc-ne-2} & 0.462 & 0.450 & 0.468 & 0.445 & 0.472 \\
        \bottomrule
      \end{tabular}
 }
  \end{subtable}
  \begin{subtable}{.49\linewidth}
    \centering
    \caption{Results of MLP models.}
    \resizebox{\linewidth}{!}{
      \begin{tabular}{llllll}
        \toprule
        & Last mean & All-levels mean & Last median & All-levels median & p value \\
        \midrule
        \texttt{WLn-1} & 0.192 & 0.212 & 0.218 & 0.229 & 0.969 \\
        \texttt{WLn-2} & 0.204 & 0.201 & 0.252 & 0.229 & 0.826 \\
        \texttt{WLne-1} & 0.183 & 0.182 & 0.225 & 0.217 & 0.811 \\
        \texttt{WLne-2} & 0.236 & 0.187 & 0.289 & 0.237 & 0.237 \\
        \texttt{WLc-n-1} & 0.394 & 0.397 & 0.412 & 0.420 & 0.910 \\
        \texttt{WLc-n-2} & 0.375 & 0.374 & 0.381 & 0.386 & 0.695 \\
        \texttt{WLc-ne-1} & 0.381 & 0.402 & 0.378 & 0.426 & 0.534 \\
        \texttt{WLc-ne-2} & 0.394 & 0.391 & 0.417 & 0.418 & 0.826 \\
        \bottomrule
      \end{tabular}
 }
  \end{subtable}
  \caption{Results of the ML models trained for Borda count maximisation on \texttt{WL} features with (all-levels) and without (last) all aggregation levels. Results report mean and median scores, as well as the p-value calculated with the Wilcoxon test.}
  \label{tab:all-levels-borda}
\end{table}

\begin{table}[t]
  \centering
  \begin{subtable}{.49\linewidth}
    \centering
    \caption{Results of SVM models.}
    \resizebox{\linewidth}{!}{
      \begin{tabular}{llllll}
        \toprule
        & Last mean & All-levels mean & Last median & All-levels median & p value \\
        \midrule
        \texttt{WLn-1} & 0.589 & 0.596 & 0.591 & 0.600 & 0.364 \\
        \texttt{WLn-2} & 0.607 & 0.591 & 0.603 & 0.603 & 0.217 \\
        \texttt{WLne-1} & 0.589 & 0.597 & 0.600 & 0.596 & 0.336 \\
        \texttt{WLne-2} & 0.605 & 0.604 & 0.603 & 0.603 & 0.974 \\
        \texttt{WLc-n-1} & 0.695 & 0.663 & 0.681 & 0.655 & 0.030 \\
        \texttt{WLc-n-2} & 0.687 & 0.664 & 0.690 & 0.672 & 0.089 \\
        \texttt{WLc-ne-1} & 0.693 & 0.674 & 0.690 & 0.664 & 0.251 \\
        \texttt{WLc-ne-2} & 0.690 & 0.659 & 0.690 & 0.661 & 0.060 \\
        \bottomrule
      \end{tabular}
 }
  \end{subtable}

  \vspace{1em} 

  \begin{subtable}{.49\linewidth}
    \centering
    \caption{Results of RF models.}
    \resizebox{\linewidth}{!}{
      \begin{tabular}{llllll}
        \toprule
        & Last mean & All-levels mean & Last median & All-levels median & p value \\
        \midrule
        \texttt{WLn-1} & 0.671 & 0.678 & 0.672 & 0.672 & 0.584 \\
        \texttt{WLn-2} & 0.671 & 0.677 & 0.672 & 0.672 & 0.611 \\
        \texttt{WLne-1} & 0.668 & 0.674 & 0.672 & 0.672 & 0.804 \\
        \texttt{WLne-2} & 0.675 & 0.672 & 0.672 & 0.672 & 0.866 \\
        \texttt{WLc-n-1} & 0.666 & 0.669 & 0.652 & 0.655 & 0.615 \\
        \texttt{WLc-n-2} & 0.666 & 0.661 & 0.655 & 0.655 & 0.619 \\
        \texttt{WLc-ne-1} & 0.661 & 0.665 & 0.655 & 0.661 & 0.930 \\
        \texttt{WLc-ne-2} & 0.668 & 0.659 & 0.672 & 0.655 & 0.322 \\
        \bottomrule
      \end{tabular}
 }
  \end{subtable}
  \begin{subtable}{.49\linewidth}
    \centering
    \caption{Results of MLP models.}
    \resizebox{\linewidth}{!}{
      \begin{tabular}{llllll}
        \toprule
        & Last mean & All-levels mean & Last median & All-levels median & p value \\
        \midrule
        \texttt{WLn-1} & 0.581 & 0.577 & 0.586 & 0.586 & 0.841 \\
        \texttt{WLn-2} & 0.596 & 0.594 & 0.586 & 0.603 & 0.714 \\
        \texttt{WLne-1} & 0.574 & 0.581 & 0.574 & 0.583 & 0.365 \\
        \texttt{WLne-2} & 0.594 & 0.586 & 0.591 & 0.603 & 0.532 \\
        \texttt{WLc-n-1} & 0.631 & 0.633 & 0.638 & 0.621 & 0.828 \\
        \texttt{WLc-n-2} & 0.616 & 0.616 & 0.621 & 0.621 & 0.941 \\
        \texttt{WLc-ne-1} & 0.630 & 0.630 & 0.632 & 0.626 & 0.935 \\
        \texttt{WLc-ne-2} & 0.627 & 0.616 & 0.621 & 0.621 & 0.571 \\
        \bottomrule
      \end{tabular}
 }
  \end{subtable}
  \caption{Results of the ML models trained for algorithm selection accuracy maximisation on \texttt{WL} features with (all-levels) and without (last) all aggregation levels. Results report mean and median scores, as well as the p-value calculated with the Wilcoxon test.}
  \label{tab:all-levels-accuracy}
\end{table}

\begin{table}[t]
  \centering
  \begin{subtable}{.49\linewidth}
    \centering
    \caption{Results of SVM models.}
    \resizebox{\linewidth}{!}{
      \begin{tabular}{llllll}
        \toprule
        & Last mean & All-levels mean & Last median & All-levels median & p value \\
        \midrule
        \texttt{WLc-n-1} & 0.622 & 0.620 & 0.633 & 0.617 & 0.935 \\
        \texttt{WLc-n-2} & 0.611 & 0.626 & 0.605 & 0.617 & 0.327 \\
        \texttt{WLc-ne-1} & 0.618 & 0.627 & 0.633 & 0.622 & 0.295 \\
        \texttt{WLc-ne-2} & 0.625 & 0.628 & 0.622 & 0.625 & 0.851 \\
        \texttt{WLn-1} & 0.625 & 0.619 & 0.617 & 0.633 & 0.615 \\
        \texttt{WLn-2} & 0.604 & 0.615 & 0.600 & 0.617 & 0.438 \\
        \texttt{WLne-1} & 0.633 & 0.610 & 0.627 & 0.617 & 0.095 \\
        \texttt{WLne-2} & 0.604 & 0.601 & 0.600 & 0.600 & 0.845 \\
        \bottomrule
      \end{tabular}
 }
  \end{subtable}

  \vspace{1em} 

  \begin{subtable}{.49\linewidth}
    \centering
    \caption{Results of RF models.}
    \resizebox{\linewidth}{!}{
      \begin{tabular}{llllll}
        \toprule
        & Last mean & All-levels mean & Last median & All-levels median & p value \\
        \midrule
        \texttt{WLc-n-1} & 0.642 & 0.651 & 0.633 & 0.650 & 0.409 \\
        \texttt{WLc-n-2} & 0.634 & 0.638 & 0.650 & 0.650 & 0.658 \\
        \texttt{WLc-ne-2} & 0.643 & 0.649 & 0.633 & 0.661 & 0.619 \\
        \texttt{WLc-ne-1} & 0.654 & 0.645 & 0.650 & 0.633 & 0.435 \\
        \texttt{WLn-1} & 0.635 & 0.645 & 0.650 & 0.650 & 0.369 \\
        \texttt{WLn-2} & 0.645 & 0.650 & 0.644 & 0.650 & 0.476 \\
        \texttt{WLne-1} & 0.643 & 0.634 & 0.647 & 0.633 & 0.464 \\
        \texttt{WLne-2} & 0.642 & 0.648 & 0.650 & 0.650 & 0.631 \\
        \bottomrule
      \end{tabular}
 }
  \end{subtable}
  \begin{subtable}{.49\linewidth}
    \centering
    \caption{Results of MLP models.}
    \resizebox{\linewidth}{!}{
      \begin{tabular}{llllll}
        \toprule
        & Last mean & All-levels mean & Last median & All-levels median & p value \\
        \midrule
        \texttt{WLn-1} & 0.621 & 0.627 & 0.633 & 0.633 & 0.975 \\
        \texttt{WLn-2} & 0.622 & 0.606 & 0.630 & 0.608 & 0.097 \\
        \texttt{WLne-1} & 0.618 & 0.620 & 0.622 & 0.633 & 0.854 \\
        \texttt{WLne-2} & 0.618 & 0.616 & 0.630 & 0.613 & 0.952 \\
        \texttt{WLc-n-1} & 0.644 & 0.645 & 0.633 & 0.644 & 0.886 \\
        \texttt{WLc-n-2} & 0.653 & 0.660 & 0.650 & 0.667 & 0.597 \\
        \texttt{WLc-ne-1} & 0.649 & 0.646 & 0.650 & 0.650 & 0.695 \\
        \texttt{WLc-ne-2} & 0.645 & 0.662 & 0.647 & 0.667 & 0.074 \\
        \bottomrule
      \end{tabular}
 }
  \end{subtable}
  \caption{Results of the ML models trained for parallel prediction on \texttt{WL} features with (all-levels) and without (last) all aggregation levels. Results report mean and median scores, as well as the p-value calculated with the Wilcoxon test.}
  \label{tab:all-levels-par}
\end{table}

\end{document}